%% file: ms.tex
\documentclass{article}
\usepackage[margin=1in]{geometry}
\newcommand\blfootnote[1]{%
  \begingroup
  \renewcommand\thefootnote{}\footnote{#1}%
  \addtocounter{footnote}{-1}%
  \endgroup
}

\RequirePackage[numbers,compress]{natbib}

\usepackage{microtype}
\usepackage{graphicx}
\usepackage{booktabs} % for professional tables

\input{files/defs-math}

\usepackage{amsmath}
\usepackage{caption}
\usepackage{subcaption}
\usepackage{float}
\usepackage{bm}
\usepackage{graphicx}
\usepackage{tabularx}
\usepackage{wrapfig}
\usepackage{enumitem}
\usepackage{hyperref}

\usepackage{algorithm}
\usepackage{algorithmic}

\usepackage{amsmath}
\usepackage{amssymb}
\usepackage{mathtools}
\usepackage{amsthm}

\usepackage[capitalize,noabbrev]{cleveref}

\theoremstyle{plain}

\theoremstyle{definition}

\theoremstyle{remark}

\usepackage[textsize=tiny]{todonotes}
\usepackage{comment}

\title{Active metric learning and classification using similarity queries}
\author{Namrata Nadagouda, Austin Xu, Mark A. Davenport}

\begin{document}
\maketitle
\blfootnote{The authors are with the School of Electrical and Computer Engineering, Georgia Institute of Technology, Atlanta, GA, 30332 USA (emails: \{namrata.nadagouda, axu77, mdav\}@gatech.edu). This work was supported, in part, by DARPA grant FA8750-19-C020 and NSF grant CCF-2107455.}

\input{files/00-abstract}

\section{Introduction}\label{sec:intro}
\input{files/01-intro}

\section{Background and related work}\label{sec:related}
\input{files/02-related_work}

\section{Unified framework and active query selection}\label{sec:methods}
\input{files/03-methods}

\section{Experiments}\label{sec:experiments}
\subsection{Deep metric learning}
\label{sec:metric_learning}
\input{files/04-embedding_learning}

\subsection{Classification}
\label{sec:classification}
\input{files/05-image_classification}

\section{Conclusion}
\label{sec:conclusion}
\input{files/06-conclusion}

\bibliographystyle{unsrtnat}
\bibliography{references}

%%%%%%%%%%%%%%%%%%%%%%%%%%%%%%%%%%%%%%%%%%%%%%%%%%%%%%%%%%%%%%%%%%%%%%%%%%%%%%%
%%%%%%%%%%%%%%%%%%%%%%%%%%%%%%%%%%%%%%%%%%%%%%%%%%%%%%%%%%%%%%%%%%%%%%%%%%%%%%%
% APPENDIX
%%%%%%%%%%%%%%%%%%%%%%%%%%%%%%%%%%%%%%%%%%%%%%%%%%%%%%%%%%%%%%%%%%%%%%%%%%%%%%%
%%%%%%%%%%%%%%%%%%%%%%%%%%%%%%%%%%%%%%%%%%%%%%%%%%%%%%%%%%%%%%%%%%%%%%%%%%%%%%%
\clearpage
\appendix
\onecolumn
\input{files/07-supp}

\end{document}

%% file: files/defs-math.tex
\usepackage{amsfonts}
\usepackage{amsmath}
\newcommand{\R}{\mathbb{R}}

\newcommand{\E}{\operatorname{\mathbb{E}}}

\newcommand{\vct}[1]{\boldsymbol{#1}}
\newcommand{\mtx}[1]{\boldsymbol{#1}}
\newcommand{\vx}{\vct{x}}
\newcommand{\vy}{\vct{y}}
\newcommand{\vz}{\vct{z}}
\newcommand{\vzero}{\vct{0}}

\newcommand{\mG}{\mtx{G}}

\newcommand{\mZ}{\mtx{Z}}

%% file: files/00-abstract.tex
\begin{abstract}
Active learning is commonly used to train label-efficient models by adaptively selecting the most informative queries. However, most active learning strategies are designed to either learn a representation of the data (e.g., embedding or metric learning) or perform well on a task (e.g., classification) on the data. However, many machine learning tasks involve a combination of both representation learning and a task-specific goal. Motivated by this, we propose a novel unified query framework that can be applied to any problem in which a key component is learning a representation of the data that reflects similarity. Our approach builds on similarity or nearest neighbor (NN) queries which seek to select samples that result in improved embeddings. The queries consist of a reference and a set of objects, with an oracle selecting the object most similar (i.e., nearest) to the reference. In order to reduce the number of solicited queries, they are chosen adaptively according to an information theoretic criterion. We demonstrate the effectiveness of the proposed strategy on two tasks -- active metric learning and active classification -- using a variety of synthetic and real world datasets. In particular, we demonstrate that actively selected NN queries outperform recently developed active triplet selection methods in a deep metric learning setting. Further, we show that in classification, actively selecting class labels can be reformulated as a process of selecting the most informative NN query, allowing direct application of our method.
\end{abstract}

%% file: files/01-intro.tex
\begin{figure*}[t]
\centering
\begin{minipage}{\textwidth}
    \centering
    \includegraphics[scale=0.12]{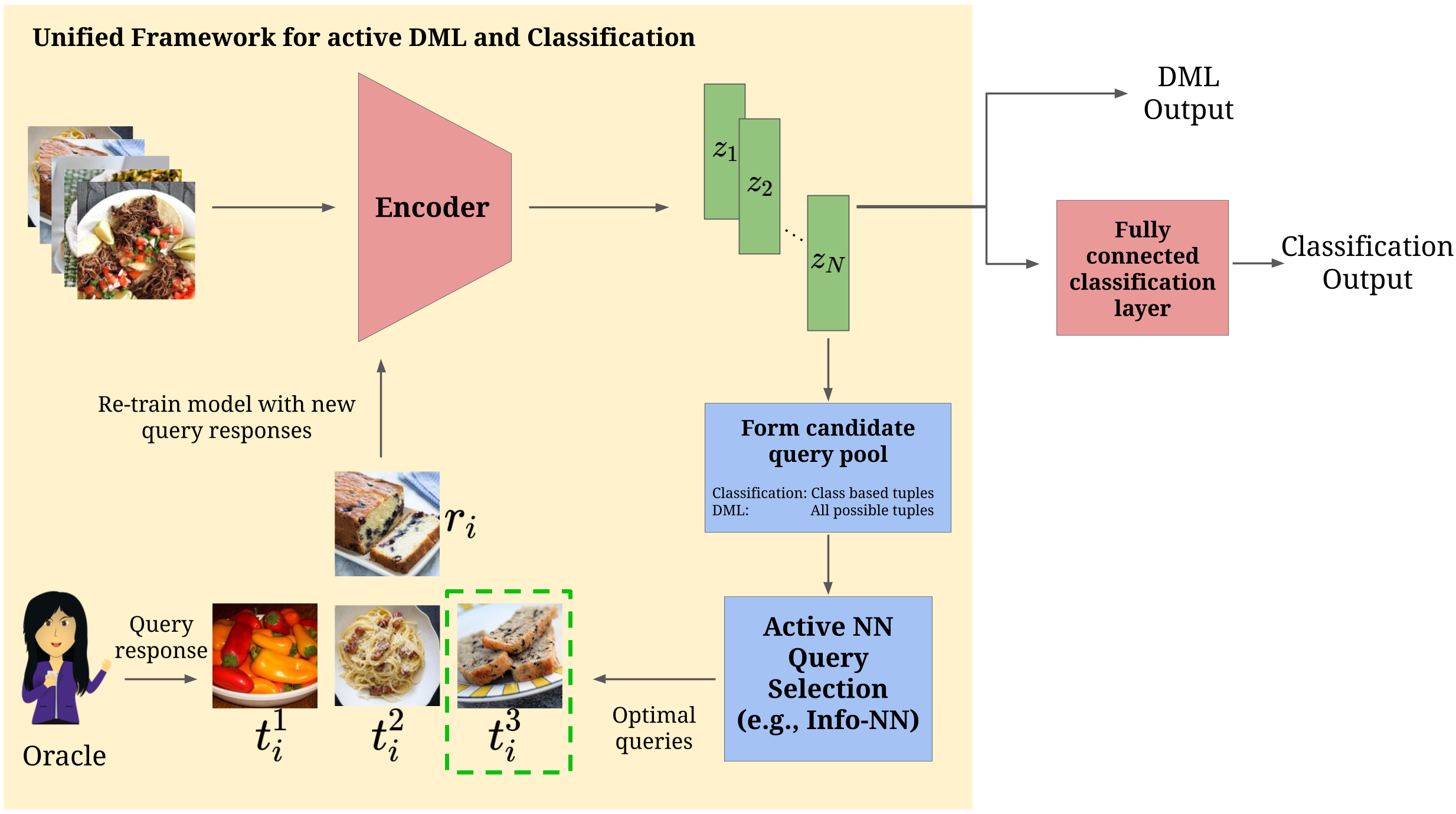}
    \vspace{-0.5\baselineskip}
    \caption{Visualization of the unified query solicitation framework with an example query. Candidate NN queries to be evaluated by the active NN query selection method are formed based on the setting (metric learning or classification). The oracle then responds to the most informative of these queries. In the case of metric learning, the response is utilized to place the reference closer to the similar item while for classification, the response is equivalent to a label corresponding to the reference (e.g., cake in this case). }
    \label{fig:unified_framework}
\end{minipage}
\vspace{-\baselineskip}
\end{figure*}

A defining feature of modern machine learning is a reliance on large volumes of human-labeled data. Perhaps the most prominent example is the existence of massive hand-labelled image datasets, but the task of acquiring large amounts of human-provided data is nearly ubiquitous in machine learning. However, such data is not free; it is often tedious and expensive to gather a sufficient number of query responses to satisfy data hungry machine learning models.

Active learning (AL) \cite{settles2009active} seeks to mitigate this issue by carefully selecting only the most informative samples to be labelled. 
%so that the models trained on these samples achieve the desired performance. 
More generally, AL attempts to identify the most informative queries to pose to an oracle. These queries can include asking for a class label or rating, or more general relational queries such as the similarity (or dissimilarity) of different items. In this paper, we focus on metric learning from perceptual similarity queries and classification, two prominent application areas for AL, and show that despite the different queries being posed to the oracle (labels in classification vs.\ similarity judgements for metric learning), there is a fundamental connection between the two problems.

%Version B
Learning an embedding or representation of the data that accurately reflects similarity between items is the goal of metric learning. Many approaches in metric learning aim to make inter-class item distances small and intra-class item distances large by using triplets of items consisting of an anchor point, a positive sample of the same class as the anchor, and a negative point of a different class \cite{hoffer2015deep}. Class labels are used as a proxy for item similarity/dissimilarity, which is only feasible if class labels are widely available. However, when given a new (unlabelled) dataset, we cannot apply this approach without manually labelling large amounts of data, and it is far from clear that class labels are the most effective mechanism for learning about similarity. We focus on one way to avoid this issue, which is to directly query an oracle for perceptual similarity information, as is done in \cite{kumari2020batch}, where triplets of the form ``Is item B or item C more similar to item A?'' are actively selected for learning an embedding of items. Active deep metric learning (DML) builds on this idea by finding the most informative queries to ask the oracle.

While seemingly dissimilar from metric learning, contemporary classification relies on models (e.g., neural networks) with the ability to learn good representations of the data from training data. Active classification focuses on how to best solicit labels for unlabelled data points, with many modern approaches either implicitly or explicitly relying on representations learned by the model to determine the most informative label. Methods that use metrics based on the predicted class probabilities, such as uncertainty \cite{gal2017deep} or consistency \cite{gao2020consistency}, implicitly rely on such representations, whereas core-set based approaches \cite{sener2018active, pinsler2019bayesian} directly use learned representations to select diverse samples. Thus, if we seek the most informative labels with respect to improving the learned representation of our classification model, the goals of active classification and active metric learning are aligned. Despite these commonalities and virtually identical learning frameworks for the two problems, to the best of our knowledge, there is no approach for query selection that is problem agnostic. In this paper, we present a unified framework, which is made feasible by a novel type of similarity query that applies to both DML and classification.

%considers a novel type of similarity query that applies to both metric learning and classification. % despite the different types of query answers needed for each problem type.

Specifically, we consider the \emph{nearest neighbor} (NN) query, which, given a reference data point $r$, asks an \emph{oracle} (e.g., a human expert) to select the most similar point from among a set of $C$ alternatives $t^1, t^2, \dots, t^C$. We denote this a length $C$ NN query. With the goal of minimizing the required number of queries, we adapt an \emph{active} query selection strategy to this query type. We take an information theoretic approach and estimate the gain in \emph{mutual information} (conditioned on previous query responses) as the criteria for selecting the most informative query, an approach that we dub \emph{Info-NN}. 

To the best of our knowledge, we are the first to study this query type. Similar ideas have been explored before, such as using UI configurations to collect multiple triplets at once \cite{wilber2014cost}, enforcing a class-similarity based quadruplet loss (one anchor, one positive point, two negative points) \cite{chen2017beyond}, and soliciting ranking queries \cite{canal2020active}. Of these approaches, \cite{chen2017beyond} is the most similar, but NN queries are 1) not confined to a particular fixed length, and 2) not restricted to using class information. The first difference allows us to generalize to any classification problem and the second allows us to collect similarity information of items of the same class, or in cases where class labels are not available.

\paragraph{Contributions.} Our main contributions are as follows. 
\begin{enumerate}[leftmargin=*, noitemsep, topsep=0pt]
    \item We propose a novel type of similarity query, called the NN query (Sec.~\ref{sec:methods}).
    \item We re-cast active classification as finding the most informative NN query, which allows us to unify active classification and active DML under one framework. This framework is flexible enough to accommodate \textit{any} active NN query selection method (Secs.~\ref{sec:classification_as_NN} and~\ref{sec:unified_framework}). 
    % \item We present two active NN query selection methods, one of which is based on a novel method procedure to compute mutual information (Sec.~\ref{sec:info-nn}).
    \item We empirically validate %\footnote{Code for all experiments is included in the supplementary material.} 
    DML and classification performance using our unified framework and novel NN query selection method (Secs.~\ref{sec:metric_learning} and ~\ref{sec:classification}).
\end{enumerate}

%% file: files/02-related_work.tex
%Active learning has been studied in a wide variety of fields ranging from natural language processing \cite{olsson2009literature}, analysis of medical images \cite{budd2019survey}, ranking systems \cite{jamieson2011active, long2014active} and more recently, in almost all fields that use deep learning models \cite{ren2020survey}. Below, we list the works relevant to the problems considered in this work.
% There are three main settings that have been commonly considered in most of these works \cite{settles2009active}: 1) membership query synthesis, 2) stream-based selective sampling and 3) pool-based sampling. The first scenario supposes that we can generate the desired queries synthetically, while the second one assumes the availability of a fixed set of queries in a streaming fashion. In this work, we consider the pool based sampling framework where we assume access to a fixed collection of queries. The unanswered queries in the pool are evaluated for their informativeness and selected accordingly for obtaining response from an oracle.  

\paragraph{Metric learning.}
Learning embeddings from similarity-based comparisons has been previously studied in a variety of scenarios \cite{agarwal2007generalized, van2012stochastic, terada2014local, amid2015multiview, kleindessner2017kernel,karaletsos2015bayesian, veit2017conditional, ma2019robust, ghosh2019landmark}, spanning everything from utilizing non-metric multidimensional scaling (MDS) to accommodating noisy/corrupted triplets to examining deeper connections to kernels. The importance of learning meaningful embeddings is shown in various applications such as face verification \cite{sankaranarayanan2016triplet}, fine-grained classification \cite{wah2014similarity}, extracting usable information from crowd-sourcing \cite{kajino2012convex}, and even fashion recommendations \cite{vasileva2018learning}. To complement these techniques, active query selection methods have been developed which examine uncertainty \cite{tamuz2011adaptively}, exploit a low-dimensionality \cite{jamieson2011low}, incorporate auxiliary features \cite{heim2015active}, and utilize Bayesian techniques \cite{lohaus2019uncertainty}. However, all of these methods are designed for non-parametric embedding techniques (e.g., MDS) which cannot easily generate a corresponding embedding given new items.

More recently, deep metric learning (DML) has aimed to overcome these limitations~\cite{kaya2019deep}. DML trains a neural network to learn an embedded representation that respects similarity information. In particular, many triplet-based DML methods assume knowledge of class labels for items, and attempt to minimize inter-class distances while maximizing intra-class distances \cite{hoffer2015deep, ge2018deep, chen2017beyond}. Although class labels may not always be available, very few works consider the case of DML with perceptual similarity queries, especially in an active manner. Recently, active similarity query selection methods for DML that focus on finding batches of non-redundant \textit{triplets} have been proposed \cite{kumari2020batch} by encouraging both informativeness (measured by entropy) and diversity (through a variety of heuristic approaches) within the selected batch. Our method adopts a similar framework as \cite{kumari2020batch}, but we utilize mutual information to find informative NN queries.

\vspace{-0.7\baselineskip}
\paragraph{Classification.}\label{sec:related_active}
Traditionally, active learning has been used with support vector machines and Gaussian processes for image classification \cite{joshi2009multi, tuia2009active, kapoor2007active, houlsby2011bayesian}. More recently, a variety of active methods based on uncertainty \cite{gal2017deep, wang2016cost, kirsch2019batchbald,  song2019combining}, diversity \cite{sener2018active, pinsler2019bayesian, kirsch2019batchbald}, and consistency \cite{gao2020consistency} have been used for training deep neural networks in the supervised and semi-supervised classification settings. In these settings, the goal is to learn a model for predicting the class probabilities on a dataset %, say $\mathcal{X} = \{\vx_i\}_{i=1}^N$ 
consisting of points belonging to $C$ classes. %, $\{y_i\}_{i=1}^N \in \{1, 2, \dots, C\}$. 
We assume access to an initial labelled %$\mathcal{L} = \{\vx_i, y_i\}_{i=1}^j$ 
and unlabelled %, $\mathcal{U} = \{\vx_i, y_i\}_{i=j+1}^N$ 
set of samples. The samples from the unlabelled pool are iteratively evaluated for informativeness and labelled accordingly. Based on feedback from the oracle, we can learn a model in either supervised (using only the labelled data) or in semi-supervised (using all data) settings.

% Uncertainty is usually computed based on the predicted class probabilities with variants such as high entropy, minimum confidence (probability associated with the predicted class) and minimum margin (difference between the probabilities associated with the top two likely classes) being used to select the samples.
Some active classification approaches \cite{houlsby2011bayesian, kirsch2019batchbald} consider mutual information between the model parameters and the predicted class probabilities to select the most informative samples, while some others \cite{sener2018active, pinsler2019bayesian} follow a coreset based approach to select a subset of diverse samples such that the model learned with these samples best approximates the one learned on the entire data. In \cite{sener2018active}, the authors use the features learned by the model to select the samples such that the maximum distance between an unlabelled sample and its nearest labelled sample is minimized. The method in \cite{pinsler2019bayesian} chooses samples such that the model posterior with the selected samples best approximates the posterior with the complete data. 

Our method derives inspiration from \cite{houlsby2011bayesian, gal2017deep, kirsch2019batchbald} in using mutual information to evaluate informativeness, but we consider mutual information between the features and the predicted class probabilities computed based on the inter-sample distances in the feature space. Our approach is similar to the work of \cite{sener2018active} in that both use the Euclidean distances of the features learned by the neural network. However, their focus is only on coverage of the entire feature space, whereas we select samples with the goal of improving the learned embedding. Apart from these, there are a few works that focus on active discriminative representation learning. In \cite{zhang2017active}, the authors propose an AL approach for text classification that selects instances containing words which are likely to most affect the embeddings by computing the \textit{expected gradient length} with respect to the embeddings. A multi-armed bandit based method that uses networking data and learned representations for adaptively labelling informative nodes is suggested in \cite{gao2018active} to learn network representations. However, to the best of our knowledge, no framework of active representation learning has been applied to image classification before and none of the above methods propose a generalized querying strategy.

% % Metric learning - problem statement
% The goal of active deep metric learning is to select the most informative similarity queries to directly ask the oracle. These queries are utilized to learn a model that transforms data features to a semantically meaningful representation of the data. We consider representations where similar items are close together and dissimilar items are far apart in Euclidean space. Given a dataset of $N$ items, we form a list of potential similarity queries to ask the oracle (e.g., triplets or nearest neighbor queries) and alternate between evaluating the informativeness of each query and, given the oracle's response, updating the model to enforce newly obtained similarity constraints.

% % Classification

% In the case of classification, the goal is to learn a model for predicting the class probabilities on a dataset, say $\mathcal{X} = \{\vx_i\}_{i=1}^N$ consisting of points belonging to $C$ classes, $\{y_i\}_{i=1}^N \in \{1, 2, \dots, C\}$. We assume access to an initial labelled, $\mathcal{L} = \{\vx_i, y_i\}_{i=1}^j$ and unlabelled, $\mathcal{U} = \{\vx_i, y_i\}_{i=j+1}^N$ set of samples. The samples from the unlabelled pool are iteratively evaluated for informativeness and labelled accordingly. Based on feedback from the oracle, we can learn a model in either supervised (using only the labelled data) or in semi-supervised (using all data) settings.  

%% file: files/03-methods.tex
\begin{figure}
    %\vspace{-5mm}
    \centering
    \includegraphics[width=0.275\textwidth]{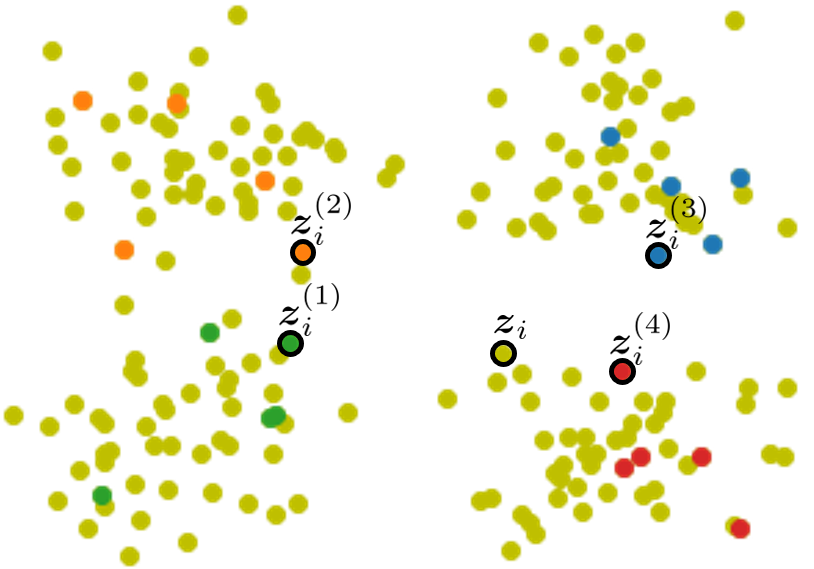}
    \caption{\small{Example of an unlabelled $\vz_i$ and the nearest labelled neighbors to $\vz_i$ from each class: $\vz_i^{(1)}, \vz_i^{(2)}, \vz_i^{(3)}, \vz_i^{(4)}$. In this example, we might expect that the most likely label would be $y_c = 4$, which could be interpreted as a nearest neighbor query response (that $\vz_i^{(4)}$ is the nearest neighbor).}} 
    \label{fig:nnquery}
\end{figure}

%\subsection{Generalized active learning framework}
In this section, we provide an overview of our proposed generalized query framework. Specifically, we show that in any classification setting where a latent representation of the data is learned, querying an oracle for a class label can be re-formulated as soliciting the oracle's feedback for an NN query, allowing us to draw the connection to metric learning. We also present Info-NN, an active method of selecting NN queries using information theoretic criterion.

Formally, a NN query $Q_i = r_i~ \cup~ T_i$ of length $C$ consists of a reference data point $r_i$ and a set of data points $T_i = \{ t_i^1, t_i^2, \dots, t_i^C \}$, from which the oracle picks the point most similar to the reference $r_i$. Let $Y_i \in \{1, 2, \dots, C\}$ be the random variable indicating the oracle's response to the $i^{th}$ query. When $Y_i = c$, this indicates that the oracle selected $t_i^c \in T_i$ as the most similar to the reference $r_i$. A visual example of the NN query can be found in Fig. \ref{fig:unified_framework}.

%\subsection{Query selection for metric learning}
%In the setting of DML, we directly ask the oracle to respond to NN queries by presenting the oracle a reference object and a set of $C$ objects, from which the oracle selects the item which is most similar to the reference object. Given a set of $N$ items, we can form a pool of potential NN queries, from which we can select the most informative queries by utilizing active learning methods, such as Info-NN. 

\subsection{Classification as a NN query selection problem}\label{sec:classification_as_NN}

We approach AL for classification as one chiefly of selecting labels that will improve the feature representation, as most modern classification techniques (e.g., neural networks) can be interpreted as learning an embedding that enables simple linear classifiers to be effective. We do this via an analogy in which obtaining the class label for an unlabelled sample is equivalent to a particular NN query response. 
   
Consider a dataset $\mathcal{X} = \{\vx_i\}_{i=1}^N$ consisting of points belonging to $C$ classes, $\{y_i\}_{i=1}^N \in \{1, 2, \dots, C\}$. We assume access to an initial labelled, $\mathcal{L} = \{\vx_i, y_i\}_{i=1}^j$ and unlabelled, $\mathcal{U} = \{\vx_i, y_i\}_{i=j+1}^N$ set of samples. Let $\mZ = \{\vz_i\}_{i=1}^N$ represent initial estimates of the embeddings for the dataset according to a model learned on an initial set of labelled samples. Now suppose we want to choose a new point $\vx_{j+1}$ from $\mathcal{U}$ whose label $y_{j+1}$ we will obtain. For any $\vx_i$ in $\mathcal{U}$, consider its embedding $\vz_i$ in the feature space and the nearest neighbor to $\vz_i$ from $\mathcal{L}$ for each class, i.e., 
\[
\vz_i^{(c)} = \mathop{\arg \min}_{\vz_\ell \in \mathcal{L}_c} \|\vz_{\ell} -  \vz_i \|_2,
\]
where $\mathcal{L}_c = \{ \vz_{\ell} : (\vx_{\ell},y_{\ell}) \in \mathcal{L}, y_{\ell} = c\}$.  An example of an unlabelled $\vz_i$ and the nearest labelled neighbors to $\vz_i$ from each class is illustrated in Fig.~\ref{fig:nnquery}. 

%Wrap fig for the informative image if needed
%\begin{comment}

%\end{comment}

Note that if the embedding that we have learned does a reasonable job of representing similarity (as it pertains to the task of classification), then we would expect that the most likely label for $\vz_i$ would correspond to the class $c$ for which $\vz_i^{(c)}$ is closest to $\vz_i$. Thus, we can interpret the label $y_i$ as a response to the nearest neighbor query in which $\vz_i$ is the reference to which $\vz_i^{(1)}, \vz_i^{(2)}, \ldots, \vz_i^{(C)}$ are compared. (For computational reasons, one may choose to not use all $C$ nearest neighbors in practice.) Because this NN query response reveals information about the relative locations of items in the learned representation, retraining the classification model with the new oracle response should improve the representation. \textbf{This is the key idea behind our approach: select NN queries (or equivalently, points to label) that result in the best improvement of the embedding.} 

\begin{algorithm}[t]
\caption{Info-NN-embedding}
\label{alg:info-nn-emb} 
\textbf{Input:} Embedding $\mZ$, candidate queries $Q$, num. samples $n_s$
%, variance $\sigma^2$
\begin{algorithmic}
    \STATE $I \leftarrow $ empty list of size $|Q|$ (Mutual information values for candidate queries)
    \STATE $p_{n}, H_{n} \leftarrow $ empty lists of size $|Q|$
    \FOR{$i = 1$ \textbf{to} $n_s$}
    \STATE $\tilde{\mZ} \leftarrow \mZ + \mG$, elements of $\mG$ drawn i.i.d from $\mathcal{N}(0, \sigma^2)$.
        \FOR{$q \in Q$}
            \STATE $r \leftarrow $ first element of $Q$
            \STATE $T \leftarrow Q \backslash \{r\}$
            \STATE $D_q \leftarrow$ distance of every item in $T$ to $r$ in $\tilde{\mZ}$
            \STATE $Y_q \leftarrow$ query response using $D_q$
            \STATE $p_{n}[q] \leftarrow p_{n}[q] + p(Y_q | D_q)$ (cumulative probability)
            \STATE $H_{n}[q] \leftarrow H_{n}[q] + H[p(Y_q | D_q)]$ (cumulative entropy)
        \ENDFOR
    \ENDFOR\\
    \FOR {$q \in Q$}
        \STATE $I[q] \leftarrow H \left[ \frac{p_{n}[q]}{n_s} \right] - \frac{H_{n}[q]}{n_s}$
    \ENDFOR
\end{algorithmic}
\textbf{Output:} $I$
\end{algorithm}

\subsection{Unified framework for active classification and metric learning}\label{sec:unified_framework}
This view of active classification gives rise to a unified framework which can be used in either active classification or active DML: from a pool of candidate NN queries, choose the most informative query to ask the oracle, then re-train the model to incorporate the newly acquired query response. \textbf{Despite each problem seemingly requiring fundamentally different oracle responses (similarity information vs.\ labels), both problems can be tackled utilizing NN queries, and thus, the same active query selection strategy.} The main difference is the pool of candidate queries. In active DML, we can query the oracle for similarity information about any set of items, whereas in active classification, the pool of candidate NN queries is restricted to queries that contain one item from every class. This pool of candidate queries is formed by setting every $\vz_i$ corresponding to an $\vx_i \in \mathcal{U}$ as the reference point, and finding (up to) $C$ nearest neighbors of differing classes. \textbf{A critical feature of this unified framework is that it does not depend on which measure of ``informativeness'' is used. This allows for a practitioner to plug-in their desired active query selection criteria without making any modifications to the framework}, as depicted in Fig. \ref{fig:unified_framework}. In our experiments, we select the queries that maximize \textit{mutual information} for both active DML and classification experiments. In particular, we utilize two methods for computing mutual information, including a novel approach dubbed Info-NN.

\subsection{Active query selection via Info-NN}\label{sec:info-nn} 
\paragraph{Observation model.} To model the oracle's response, we use a Plackett-Luce (PL) model \cite{turner2018introduction} which is an extension of the triplet model commonly used with similarity comparisons \cite{tamuz2011adaptively}: 
\begin{equation}\label{eq:prob_model}
    P(y_i = c) = \frac{ (D^2_{ic} + \mu)^{-1}}{\sum_{j=1}^C (D^2_{ij} + \mu)^{-1}}
\end{equation}
where $D_{ic}$ denotes the distance between the embeddings of $r_i$ and $t_i^c$, and $\mu$ is a parameter set by the user.  This model captures uncertainty in the oracle responses as well as uncertainty in our current estimate of the embedding (and hence distances). The parameter $\mu$ is indicative of our confidence in the distances. Note that even though we use this model in our query selection strategy, we do \emph{not}  require that query responses are generated according to the PL model.

\begin{algorithm}[t]
\caption{Info-NN-distances}
\label{alg:info-nn-dist} 
\textbf{Input:} Embedding $\mZ$, candidate queries $Q$, num. samples $n_s$
%, variance $\sigma^2$
\begin{algorithmic}
    \STATE $I \leftarrow $ empty list of size $|Q|$ (Mutual information values for candidate queries)
    \FOR{$q \in Q$}
    \STATE $r \leftarrow $ first element of $Q$
    \STATE $T \leftarrow Q \backslash \{r\}$
    \STATE $D_q \leftarrow$ distance of every item in $T$ to $r$ in $\mZ$
    \STATE $Y_q \leftarrow$ query response using $D_q$
    \STATE $D_s \leftarrow n_s$ copies of $\mathcal{N}(D_q, \sigma^2)$
    \STATE $I[q] \leftarrow H \left[ \sum\limits_{D \in D_s} \frac{\left( p(Y_q    \: | \:   D) \right)}{n_s} \right] - \sum\limits_{D \in D_s} \frac{H \left[ p(Y_q    \: | \:   D) \right]}{n_s}$
    \ENDFOR\\
\end{algorithmic}
\textbf{Output:} $I$
\end{algorithm}

\paragraph{Active query selection criteria.} 
The main idea behind our selection strategy is to select queries that are maximally informative about the embedding while avoiding ones that do not provide new information. This goal is achieved by using mutual information between the embedding and a query as a measure of the informativeness of the query.
Let $y^{n-1} = \{ y_1, y_2, \dots, y_{n-1} \}$ denote the set of all  responses obtained after $n-1$ queries. We denote $Y_n$ to be the random variable corresponding to the oracle's response to query $Q_n$. Now consider the mutual information between the embedding $\mZ$ and the response $Y_n$:
\begin{equation}\label{eq:MI_1}
I(\mZ ; Y_n   \: | \:   y^{n-1}) = H[\mZ   \: | \:   y^{n-1}] - \underset{Y_n}{\E} (H[\mZ   \: | \:   Y_n, y^{n-1}]).
\end{equation}
This quantity measures how much information the response to query $Q_n$ would provide about the embedding, conditioned on the fact that we have already acquired the responses $y^{n-1}$ to the previous queries. This is exactly what we would like to use to select informative queries, but computing this quantity in the above form is computationally expensive. To compute this in a na\"{i}ve manner we would need to find the estimate of the embedding for every possible response to the query and compute the entropies of these estimates in the high dimensional embedding space. Fortunately, using an approach similar to~\cite{houlsby2011bayesian}, we can use the symmetry of mutual information to re-write~\eqref{eq:MI_1} as
\begin{equation}\label{eq:MI_2}
I(Y_n ;\mZ   \: | \:   y^{n-1} ) = H[Y_n    \: | \:   y^{n-1} ] - \underset{\mZ}{\E} (H[Y_n    \: | \:   \mZ, y^{n-1} ]).
\end{equation}
We can now compute entropies in the response space, which is usually much smaller than the embedding space. This second form of mutual information also provides an interesting insight about the selection strategy. The first term, which denotes the entropy of the predicted response, encourages the selection of queries which are highly uncertain for the current estimate of the embedding. The second term denotes the expected entropy of the responses predicted by the individual samples from the distribution over the embedding estimate and encourages queries for which the individual samples are fairly confident. This simultaneously avoids the acquisition of redundant queries and queries for which the oracle response is likely to be uncertain.

%%%%%%%%%%%%%%%%%%%%%%%%%%%%%%%%%%%%%%%%%%%%%%%%%%%%%%%%%%%%%%%%%%%%%%%%
%%%%%%%%%%%%%%%%%%  PLACE DML EXPERIMENTS HERE TO MOVE UP A PAGE %%%%%%%
%%%%%%%%%%%%%%%%%%%%%%%%%%%%%%%%%%%%%%%%%%%%%%%%%%%%%%%%%%%%%%%%%%%%%%%%
\begin{figure*}[t]
    \hfill
    \begin{minipage}{0.27\linewidth}
        \centering
        \includegraphics[width=\textwidth]{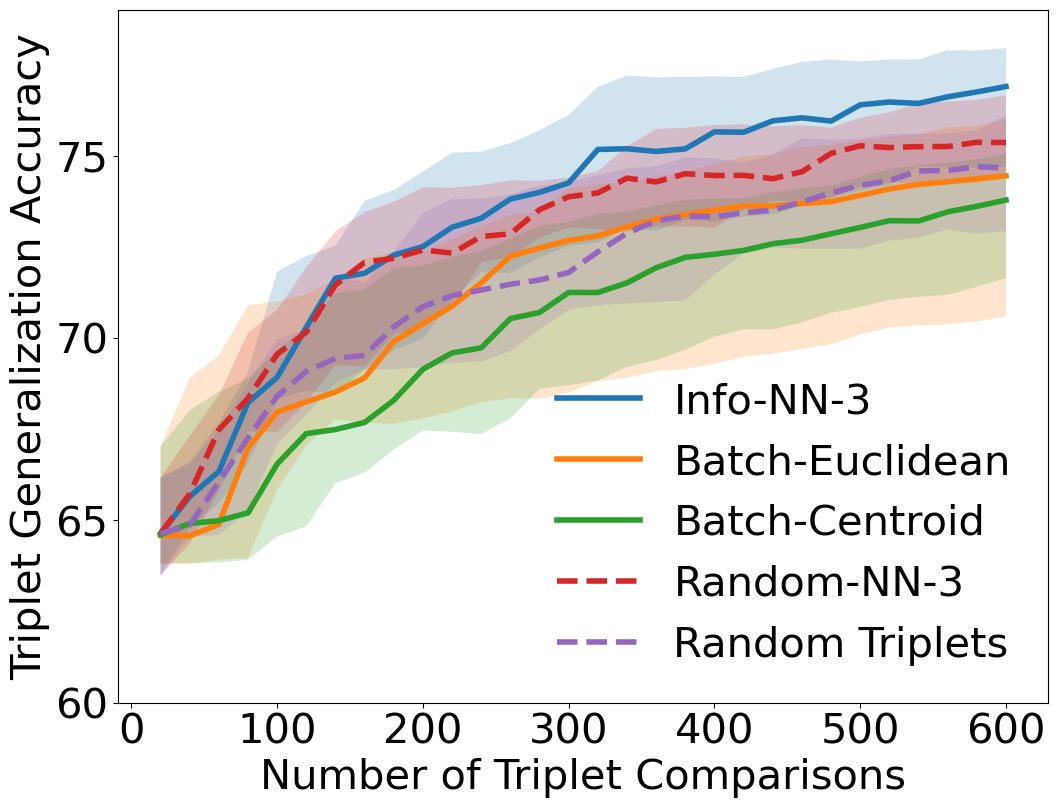}
        \label{fig:dml_synthetic}
    \end{minipage}
    \hfill
    \begin{minipage}{0.27\linewidth}
        \centering
        \includegraphics[width=\textwidth]{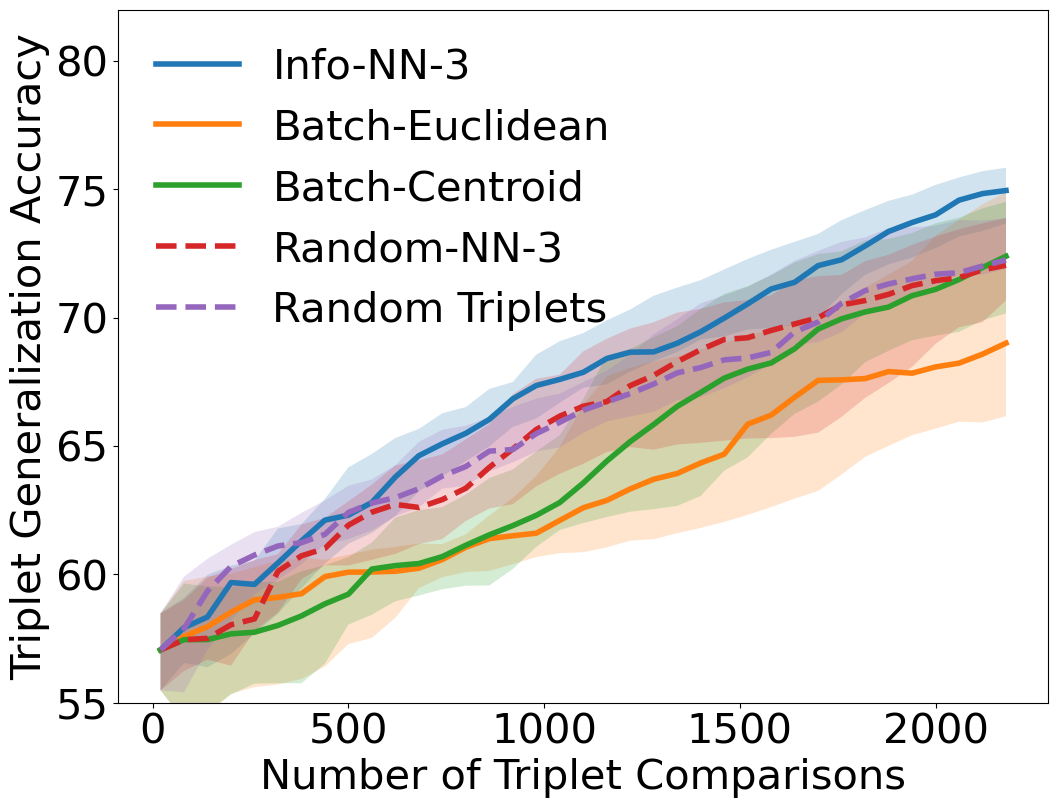}
        \label{fig:dml_food}
    \end{minipage}
    \hfill
    \begin{minipage}{0.27\linewidth}
        \centering
        \includegraphics[width=\textwidth]{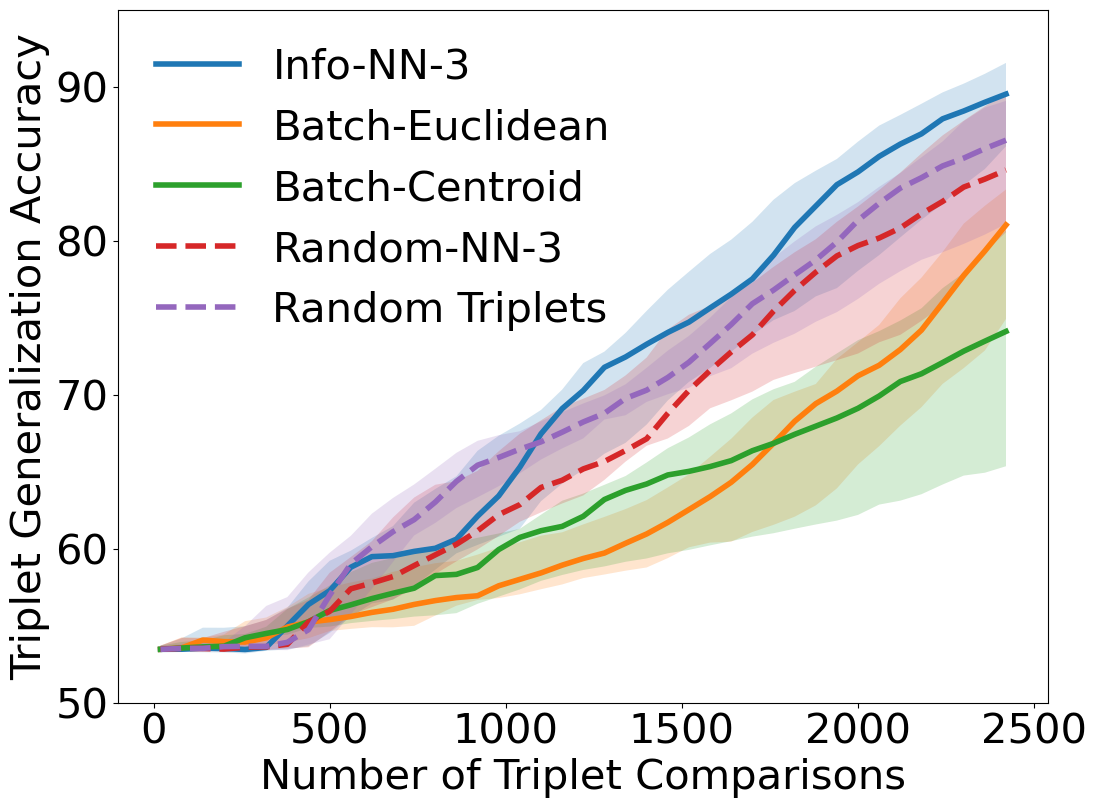}
        \label{fig:dml_admissions}
    \end{minipage}
    \hfill
    \vskip 0pt
    \vspace{-1\baselineskip}
    \hfill
    \begin{minipage}{0.27\linewidth}
        \centering
        \includegraphics[width=\textwidth]{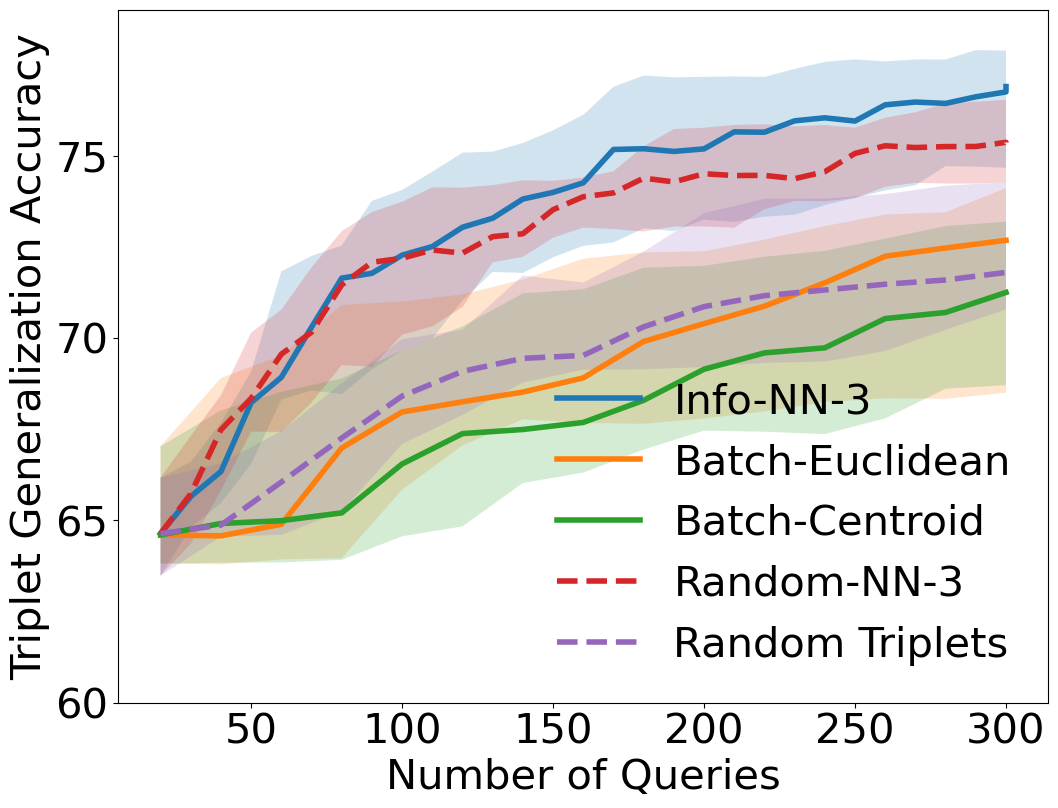}
        \label{fig:dml_synthetic_queries}
    \end{minipage}
        \hfill
    \begin{minipage}{0.27\linewidth}
        \centering
        \includegraphics[width=\textwidth]{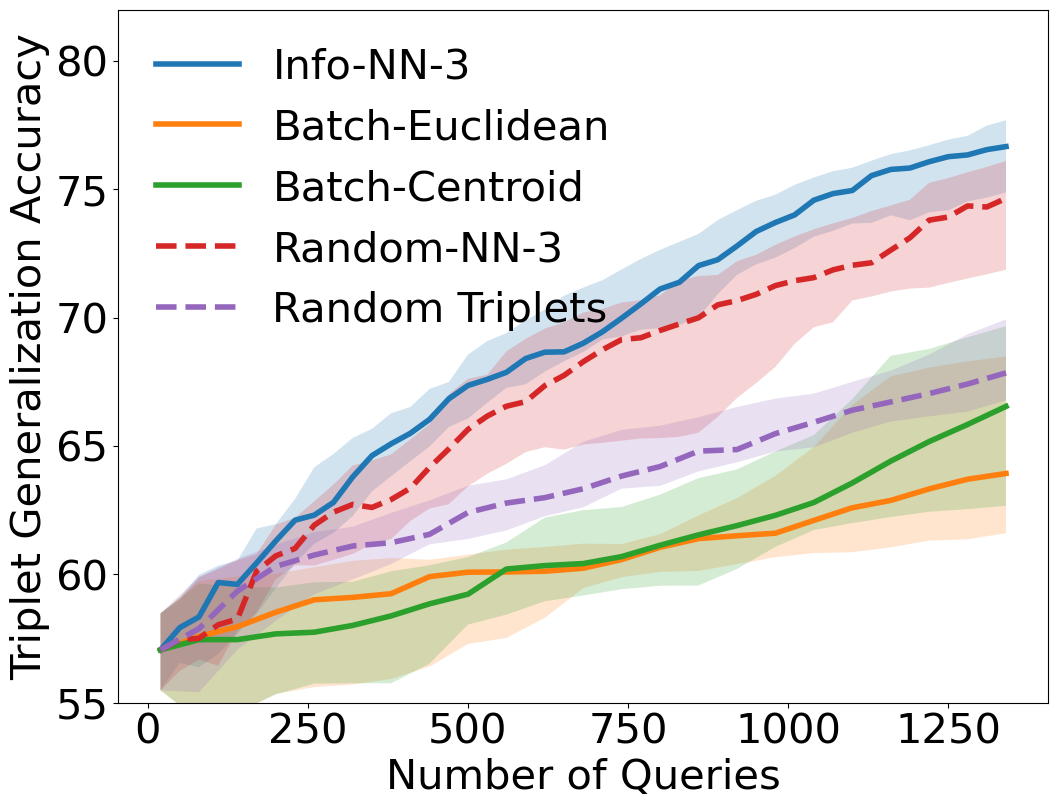}
        \label{fig:dml_food_queries}
    \end{minipage}
    \hfill
    \begin{minipage}{0.27\linewidth}
        \centering
        \includegraphics[width=\textwidth]{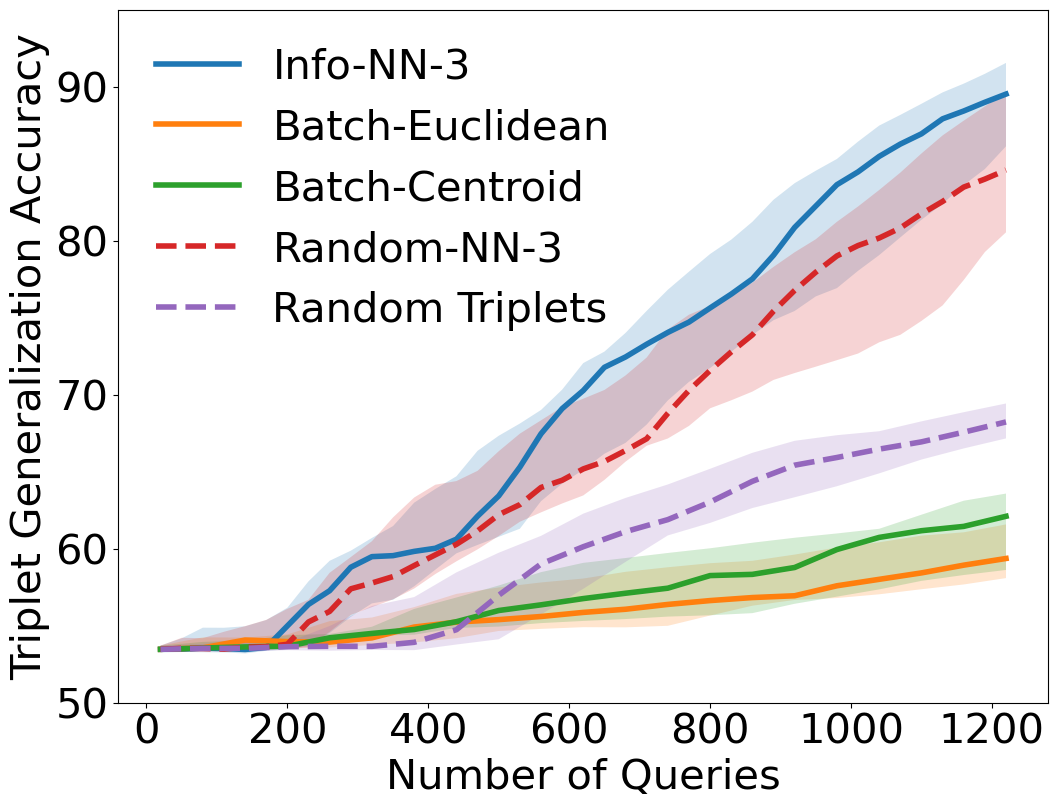}
        \label{fig:dml_admissions_queries}
    \end{minipage}
    \hfill
    \vskip 0pt
    \vspace{-1\baselineskip}
\caption{Per-triplet (top) and per-query (bottom) TGA comparison of Info-NN against active batch triplet methods and random queries on synthetic (left) and food (center), and Graduate Admission (right) datasets. Info-NN outperforms random and batch methods, and NN queries exhibit far superior per-query performance, requiring less interactions with the oracle.}
\label{fig:dml_experiments}
\vspace{-0.75\baselineskip}
\end{figure*}

\paragraph{Probabilistic inference.} Computing the mutual information as in~\eqref{eq:MI_2} requires a probabilistic estimate of the embedding. However, in many learning scenarios, posterior inference of the embedding remains computationally intractable. We utilize two Monte Carlo sampling based methods for tractable probabilistic inference. The first method, which we refer to as \textit{Info-NN-embedding}, assumes that the embedding values are normally distributed, with mean equal to the previous estimate of the embedding. With this assumption, we have a tractable means of computing the mutual information. We can further increase computational efficiency by making the stronger assumption that inter-item distances in the embedding are normally distributed, with mean equal to the previous estimates of the distances. We refer to this approach as \textit{Info-NN-distances}. In general, we use \textit{Info-NN-distances} for experiments dealing with real-world data, and \textit{Info-NN-embedding} for synthetic experiments. The two methods are presented in Alg.~\ref{alg:info-nn-emb} and Alg.~\ref{alg:info-nn-dist}, respectively, and more detailed derivations are available in the appendix.

%% file: files/04-embedding_learning.tex
\begin{figure*}[t]
\begin{minipage}{\linewidth}
    \centering
    \includegraphics[width=0.75\linewidth]{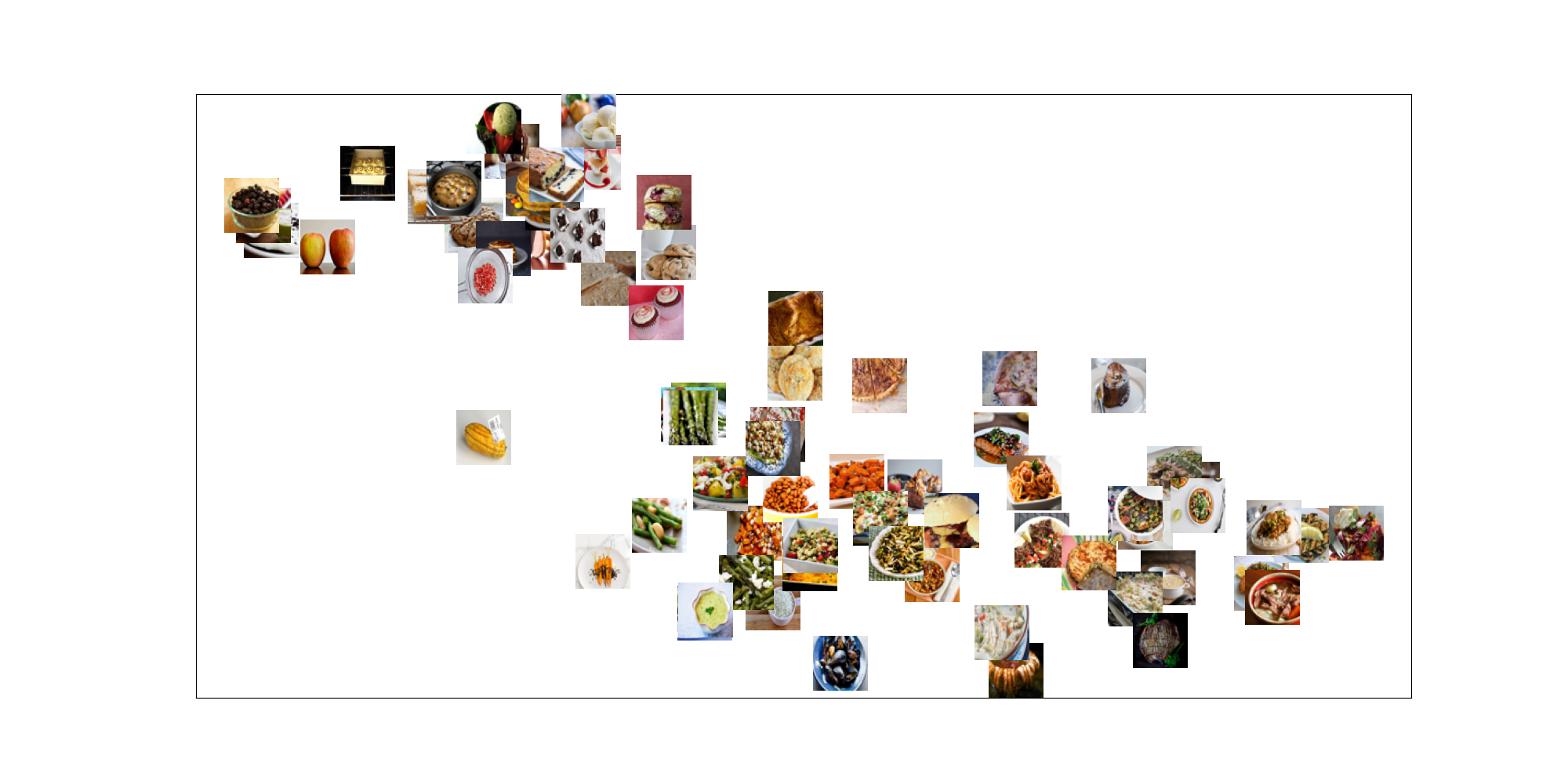}
\vspace{-2\baselineskip}
\end{minipage}
\caption{\small Visualization of food embedding learned using queries selected with Info-NN, generated using t-SNE \cite{maaten2008visualizing}. Similarly tasting objects are generally grouped together, such as vegetables (center) and fruits (top left)}
\label{fig:food_viz}
\vspace{-1\baselineskip}
\end{figure*}

%\end{comment}

\begin{figure*}[t]
    \hfill
    \begin{minipage}{0.27\linewidth}
        \centering
        \includegraphics[width=\textwidth]{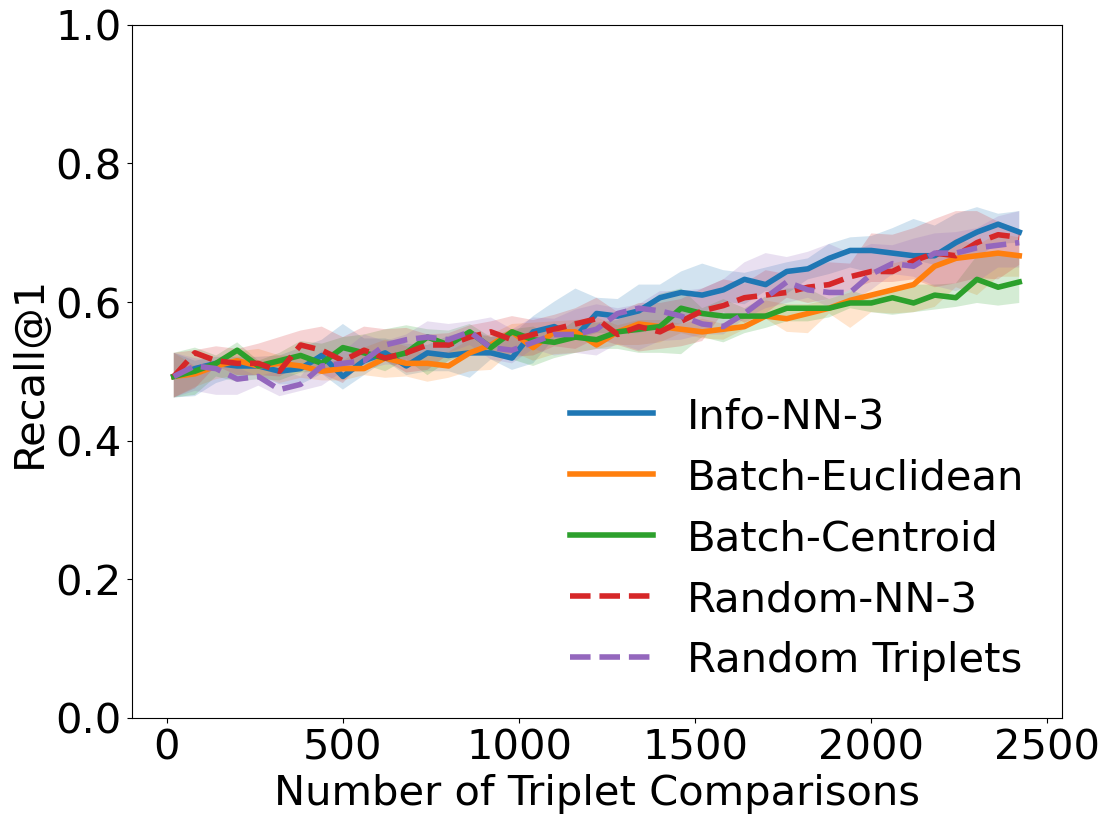}
        \label{fig:dml_adm_recall}
    \end{minipage}
    \hfill
    \begin{minipage}{0.27\linewidth}
        \centering
        \includegraphics[width=\textwidth]{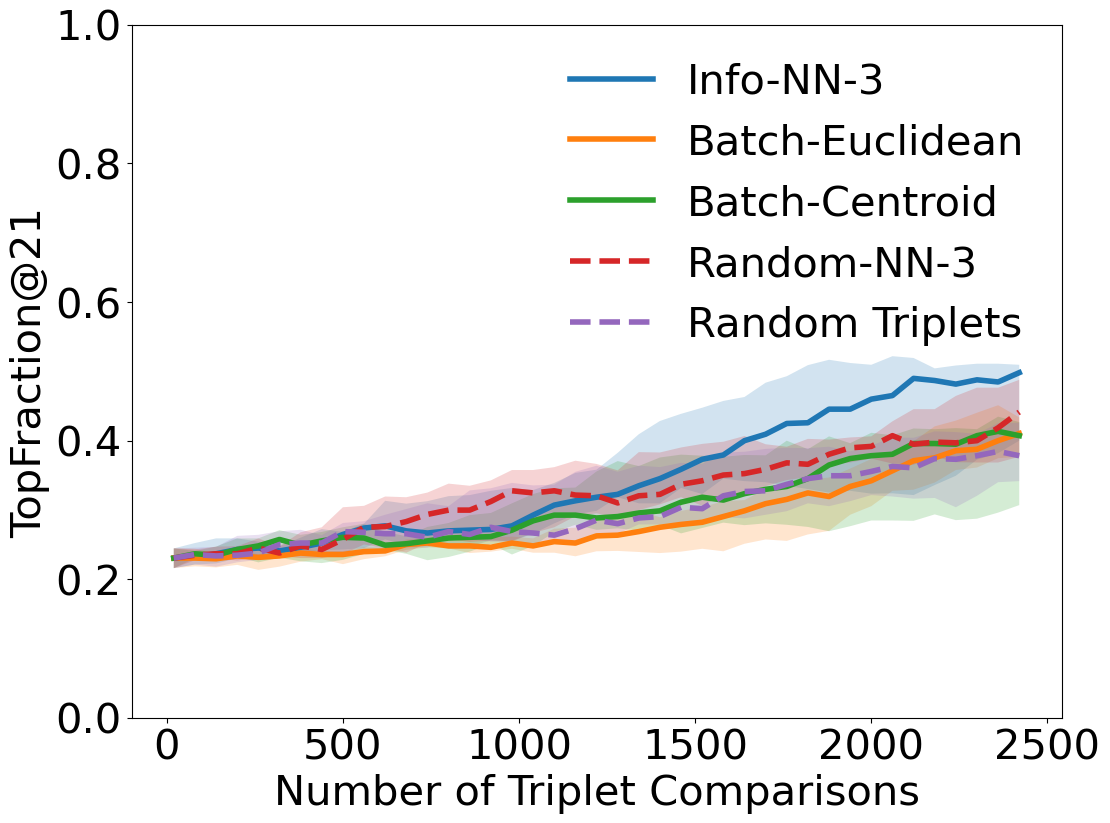}
        \label{fig:dml_adm_topk}
    \end{minipage}
    \hfill
    \begin{minipage}{0.278\linewidth}
        \centering
        \includegraphics[width=0.75\textwidth]{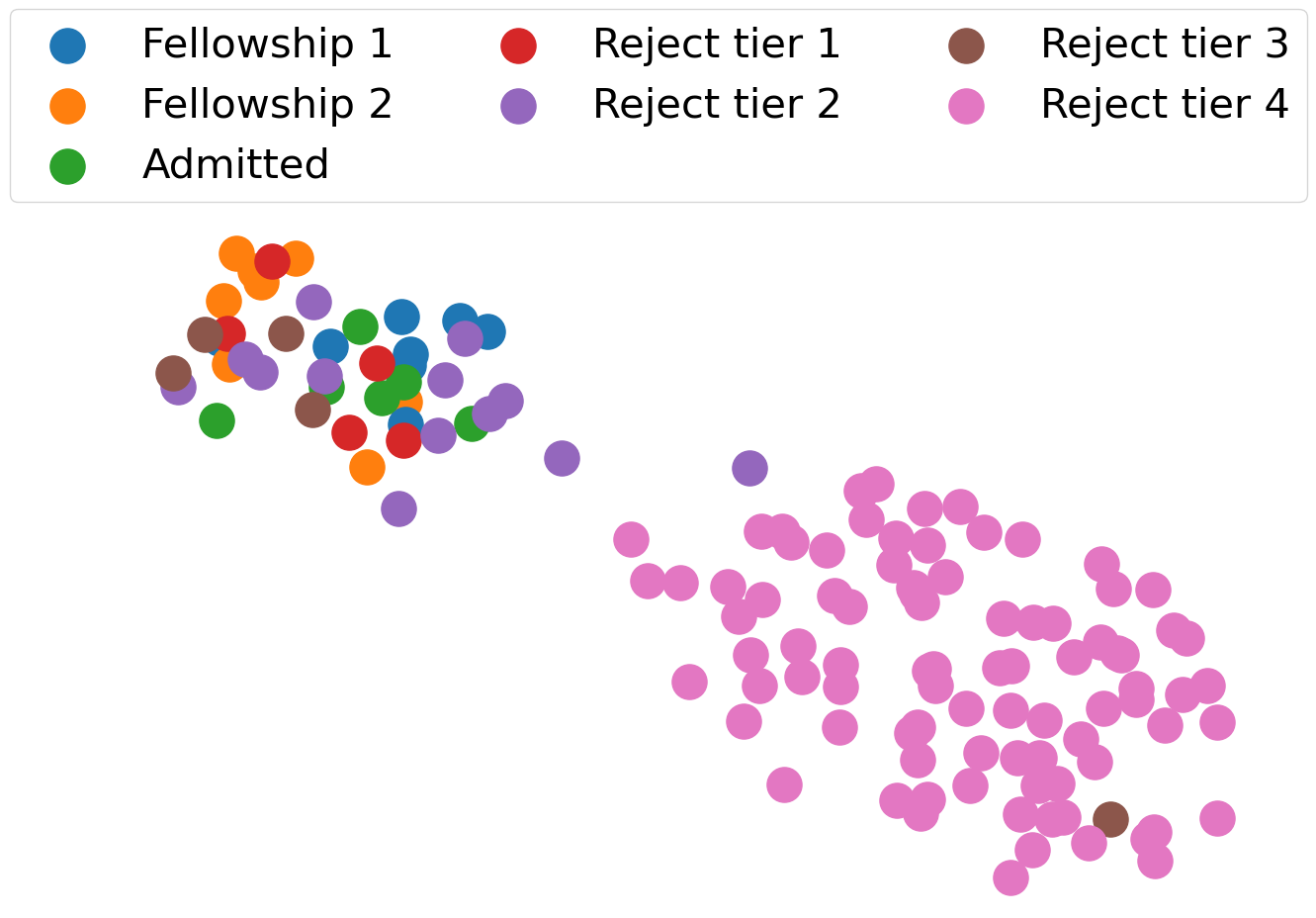}
        \label{fig:dml_adm_info_nn_vis}
    \end{minipage}
    \hfill
    \vskip 0pt
    \hfill
    \begin{minipage}{0.27\linewidth}
        \centering
        \includegraphics[width=\textwidth]{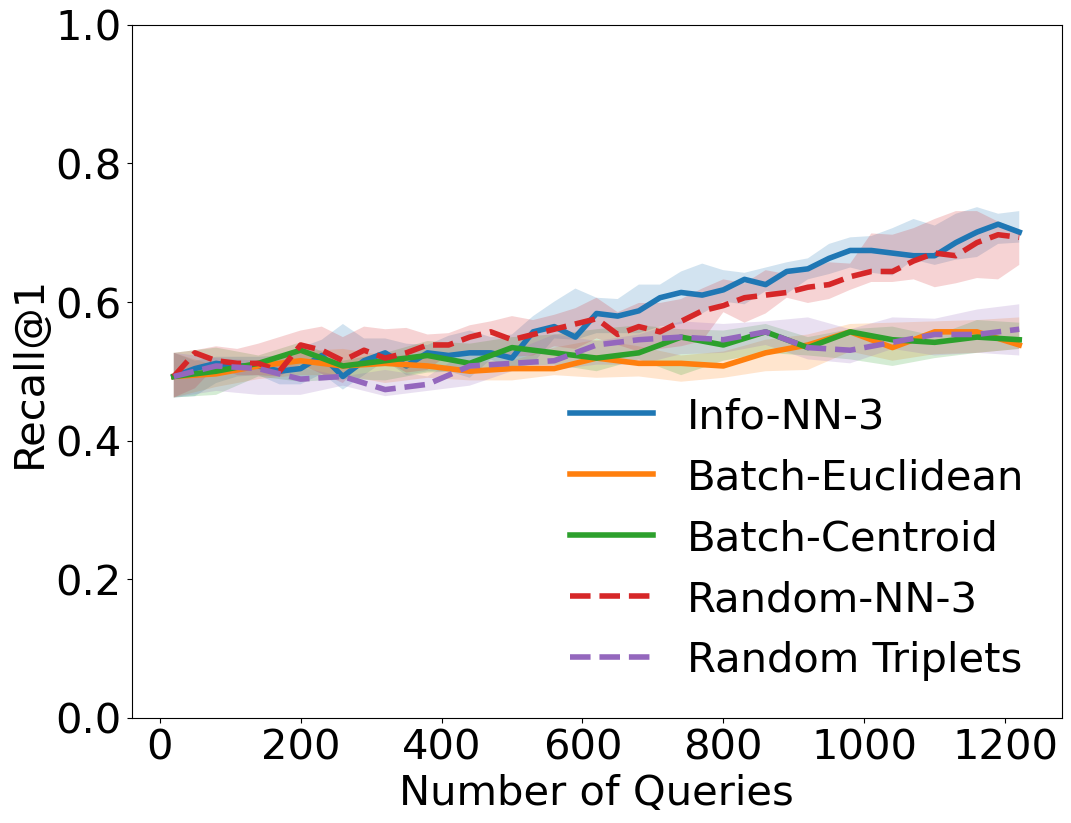}
        \label{fig:dml_adm_recall_queries}
    \end{minipage}
    \hfill
    \begin{minipage}{0.27\linewidth}
        \centering
        \includegraphics[width=\textwidth]{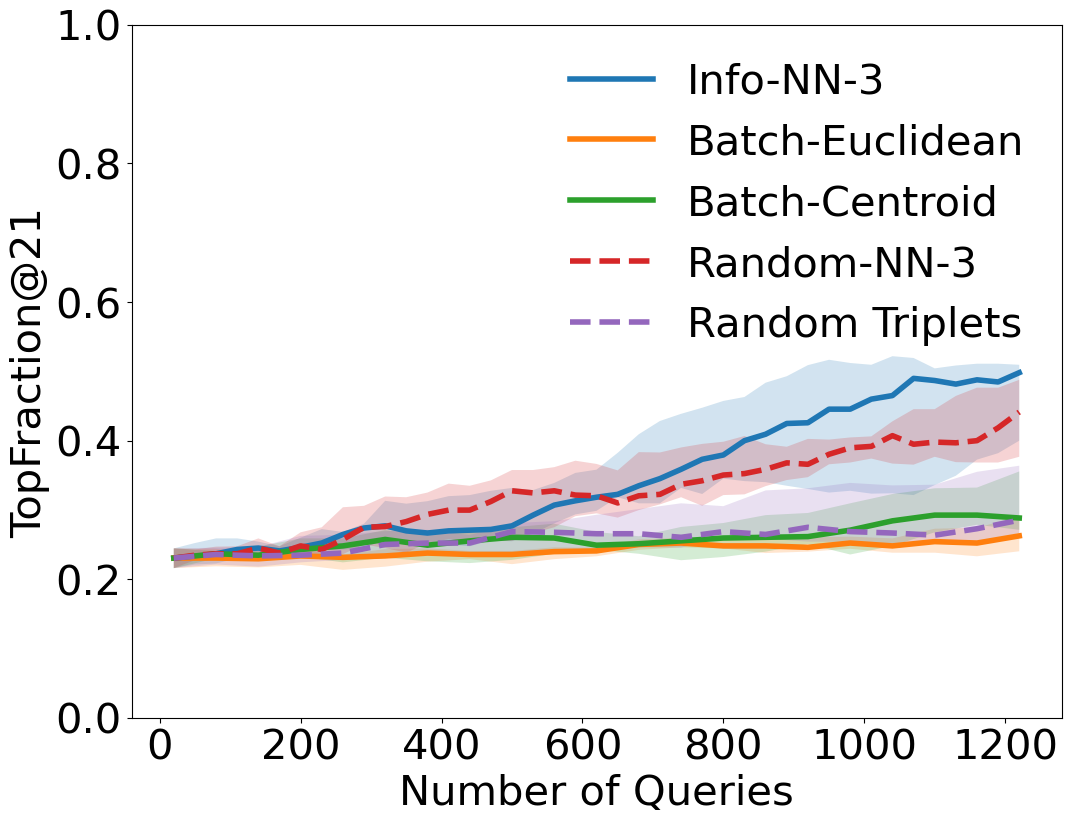}
        \label{fig:dml_adm_topk_queries}
    \end{minipage}
    \hfill
    \begin{minipage}{0.278\linewidth}
        \centering
        \includegraphics[width=0.75\textwidth]{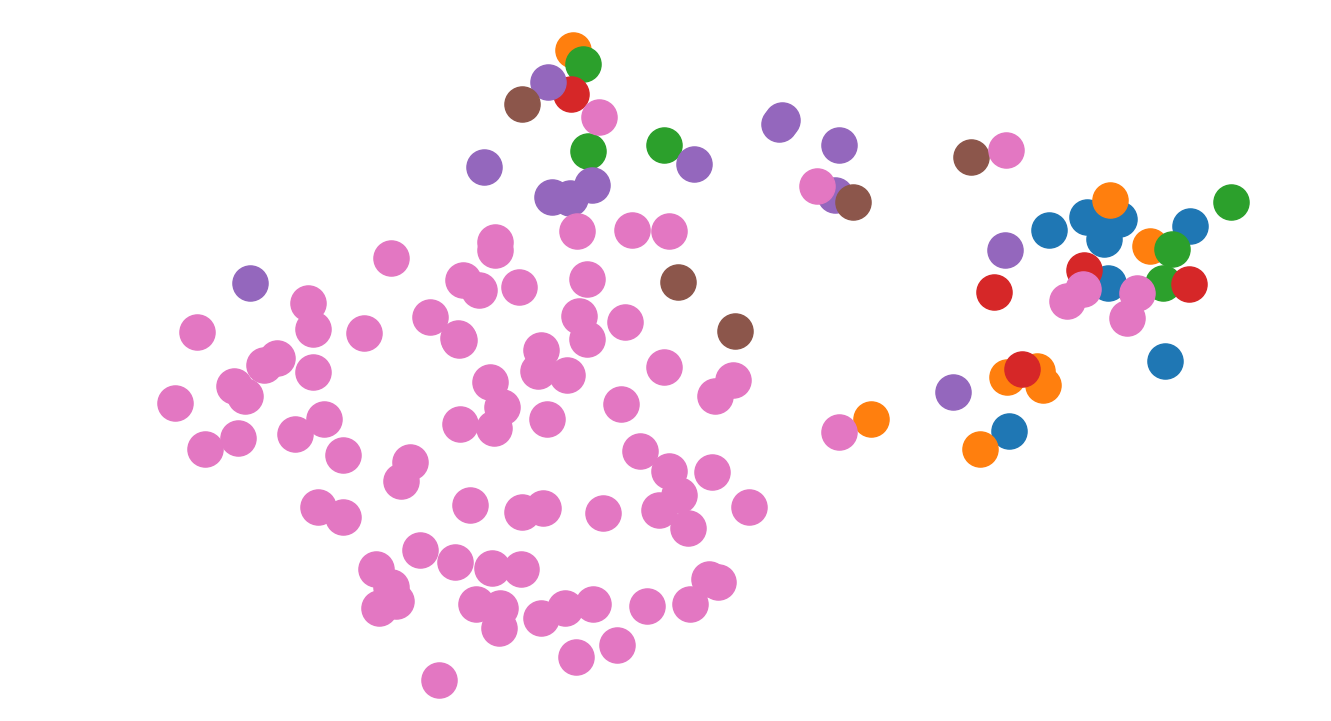}
        \label{fig:dml_adm_cent_vis}
    \end{minipage}
    \hfill
    \vskip 0pt
    \vspace{-1\baselineskip}
\caption{\small Per-triplet (top) and per-query (bottom) comparison for Info-NN against other methods Recall@$1$ (left) and TopFraction@$21$ (center). NN queries result in objects of the same class to be more nearby and group admitted students together, with Info-NN exhibiting the best performance of all methods tested. Visualization of embedding learned using Info-NN (right-top) and Batch-Centroid (right-bottom), generated using t-SNE \cite{maaten2008visualizing}. Info-NN groups more highly rated candidates closer together.}
\label{fig:dml_recall_topk_vis}
\vspace{-1\baselineskip}
\end{figure*}

In this section, we directly query an oracle with NN queries and learn a similarity embedding from query responses using a Deep Metric Learning (DML) framework.

\paragraph{Active embedding framework.}
We utilize a neural network to learn an embedding that matches the oracle's responses to similarity queries. Because a length $C$ NN query can be decomposed into $C-1$ triplets, we utilize a triplet loss \cite{weinberger2006distance}. We initialize our network with a random batch of triplets, then select batches of $B$ queries, receive oracle responses to the selected queries, add the new queries to the pool of already answered queries, and re-train our network for $100$ epochs using all prior query responses. For each experiment, we select a pool of $20,000$ training length-3 NN queries and $20,000$ testing length-3 NN queries from the set of all possible queries (decomposing NN queries into triplets for triplet based methods). 

In scenarios where re-training the network many times is computationally expensive, batch methods that select multiple queries at once are preferable. We compare the performance of Info-NN to recently developed triplet batch methods \cite{kumari2020batch}. While Info-NN can identify informative queries, batches of the most informative queries at a fixed instance may result in poor diversity of queries, as the most informative queries often cover the same areas of the space. Therefore, we utilize a very simple batch extension for DML experiments. For a batch of $B$ queries, we select the top $B^\prime \leq B$ most informative queries, then select $B - B^\prime$ queries uniformly at random from the query pool. We show that simply augmenting randomly selected queries with a set of the most informative queries can outperform methods designed specifically for batch query selection.

In our experiments, \emph{Info-NN-C} means the batch variant of Info-NN described above was used to select NN queries of length $C$, while \emph{Batch-Euclidean/Centroid} indicate methods proposed in \cite{kumari2020batch}. Finally, \emph{Random} means the query type (NN or triplet) was constructed by selecting queries uniformly at random from the training set. Precise experimental details can be found in the appendix.

\paragraph{Datasets and evaluation metrics.}
We test our active embedding technique on a variety of datasets:
\begin{itemize}[leftmargin=*,noitemsep, topsep=0pt]
    \item \textbf{Synthetic Mahalanobis Dataset:} We generate $N = 100$ items of dimension $D = 10$ from a standard normal distribution. The oracle makes perception judgements based on some randomly generated Mahalanobis metric. We introduce artificial noise by corrupting 25\% of all training queries to assess the robustness of our embedding method. We collect batches of size $B = 10$. Info-NN-embedding is used in these experiments.
    \item \textbf{Food73 Dataset:} This dataset contains $72,148$ crowdsourced triplets gathered for $73$ different food items~\cite{wilber2014cost}. We utilize $6$ L1 normalized features (bitterness, saltiness, savoriness, spiciness, sourness, and sweetness) for each food item and form $1,047,251$ length $3$ NN queries from the collected triplets. The collected triplets, and, as a result, the formed NN queries contain inconsistencies. We collect batches of size $B = 30$. Info-NN-distances is used in these experiments.
    \item \textbf{Ranked Graduate Admissions Dataset:} We obtained partially ranked lists of 133 Ph.D. applicants to [\textit{redacted for review}]. The top 22 candidates were accepted for admission, with the top 18 candidates individually ranked and the the rest of the candidates sorted into 5 tiers of varying sizes. Candidates fall into one of $7$ classes: Admitted with fellowship 1, admitted with fellowship 2, admitted without fellowship, reject (sorted into 4 tiers). For each candidate, we have access to $25$ features including GPA, letters of recommendation scores, and external fellowship application status. We form triplets across among the ranked candidates and between candidates of different tiers, resulting in $434,470$ triplets and $21,634,487$ length $3$ NN queries, and randomly corrupt 25\% of all queries to assess robustness. We collect batches of size $B = 30$. Info-NN-distances is used in these experiments.
\end{itemize}

To measure the performance of our embedding learning algorithm, we use \textit{triplet generalization accuracy}, which records the fraction of test triplets whose ordering is consistent with the learned embedding. Furthermore, for the Graduate Admissions Dataset, because we have access to class labels, we record Recall@$K$. Furthermore, to get a sense of how the algorithms group the admitted students, we compute TopFraction@K, which denotes the fraction of the $K$ nearest neighbors of the top $22$ (admitted) students that are admitted students. Because NN queries can be decomposed into triplets, we compare performance against triplet-based methods on both a \textit{per-triplet} basis and a \textit{per-query} basis (number of queries posed to the oracle). We report the median and 25\% and 75\% quantile over $20$ (synthetic), $10$ (food), and $10$ (admissions) trials.

\paragraph{Experiment results.}
As seen in Fig. \ref{fig:dml_experiments}, both versions of Info-NN are able to outperform recent methods developed specifically for batch query selection on both synthetic and challenging real-world datasets on both a per-triplet and a per-query basis. \textbf{This also demonstrates the flexibility of the unified framework; multiple active query selection methods can be plugged into the framework with consistently strong performance.} Furthermore, there seems to be minimal performance difference in selecting random NN queries vs.\ random triplets on a per-triplet basis, but using NN queries requires far fewer interactions with the oracle. From these experiments, it appears that the methods in \cite{kumari2020batch} require more of a ``warm up'' to catch up to random query performance, whereas Info-NN can consistently outperform random. Inspecting the visualization in Fig. \ref{fig:food_viz} of the learned food embedding also reveals that the embedding learned using Info-NN nicely separates savory foods from sweet foods, and can even group together similar foods, such as vegetables and fruits. Beyond triplet generalization accuracy, we can see in Fig.~\ref{fig:dml_recall_topk_vis} that Info-NN is able to outperform the same methods on both a per-triplet and per-query basis in Recall@$1$ and TopFraction@$21$, which suggests that Info-NN is more capable of grouping admitted students together. This can be visualized in Fig.~\ref{fig:dml_recall_topk_vis}, where top-rated students are more clearly grouped together in the embedding learned using Info-NN compared to the embedding learned with Batch-Centroid. Results for varying values of $K$ for Recall@$K$ and TopFraction@$K$ can be found in the appendix. %We also note that Info-NN can be utilized in non-DML settings, such as using MDS \cite{tamuz2011adaptively}, and performs on par with more complex ranking queries (see appendix).

\begin{comment}
\begin{figure*}[t]
    \hfill
    \begin{minipage}{0.4\linewidth}
        \centering
        \includegraphics[width=\textwidth]{figures/dml/adm_info_nn_vis.png}
        \label{fig:dml_adm_info_nn_vis}
    \end{minipage}
    \hfill
    \begin{minipage}{0.4\linewidth}
        \centering
        \includegraphics[width=\textwidth]{figures/dml/adm_centroid_vis.png}
        \label{fig:dml_adm_cent_vis}
    \end{minipage}
    \hfill
\caption{\small Visualization of embedding learned using Info-NN (left) and Batch-Centroid (Right), generated using t-SNE \cite{maaten2008visualizing}. The embedding learned with Info-NN groups more highly rated candidates closer together.} \vspace{-2mm}
\label{fig:dml_adm_vis}
\end{figure*}
\end{comment}

%% file: files/05-image_classification.tex
We perform experiments on active image classification in a supervised setting, using NN queries to acquire labels iteratively. Info-NN-distances is used in these experiments.

\paragraph{Label selection and experimental framework.}
To select samples using Info-NN, for every unlabelled sample, we form the corresponding nearest neighbor query and compute an estimate of the information gain provided by that query. We then request a label for the unlabelled sample corresponding to the most informative query. A simple batch extension of our query acquisition strategy, which performs a clustering of the unlabelled samples in the embedding space and selects the most informative samples from every cluster, is used in the experiments. \emph{Info-NN-C} means the batch variant of Info-NN was used to select NN queries of length $C$. We use Euclidean distances between the features learned by the last hidden layer to compute distances for the probability model. We experimented with the length of the queries and illustrate plots for the best performing values. We plot the median of the accuracy values along with the 25\% and 75\% quantiles over 3 trials. More details can be found in the appendix. % regarding the models, baselines and optimization techniques used can be found in the supplementary material.

\begin{figure*}[t]
    \hfill
    \begin{minipage}{0.27\linewidth}
        \centering
        \includegraphics[width=\textwidth]{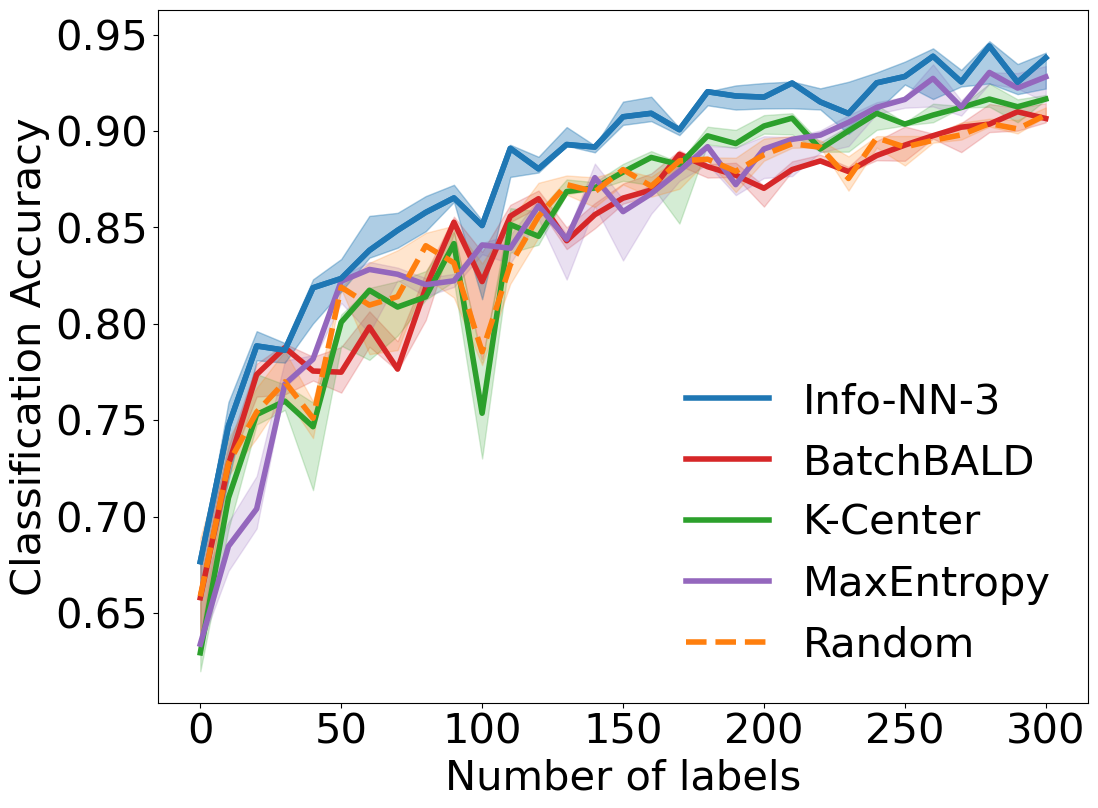}
        \label{fig:mnist_baselines}
    \end{minipage}
    \hfill
    \begin{minipage}{0.27\linewidth}
        \centering
        \includegraphics[width=\textwidth]{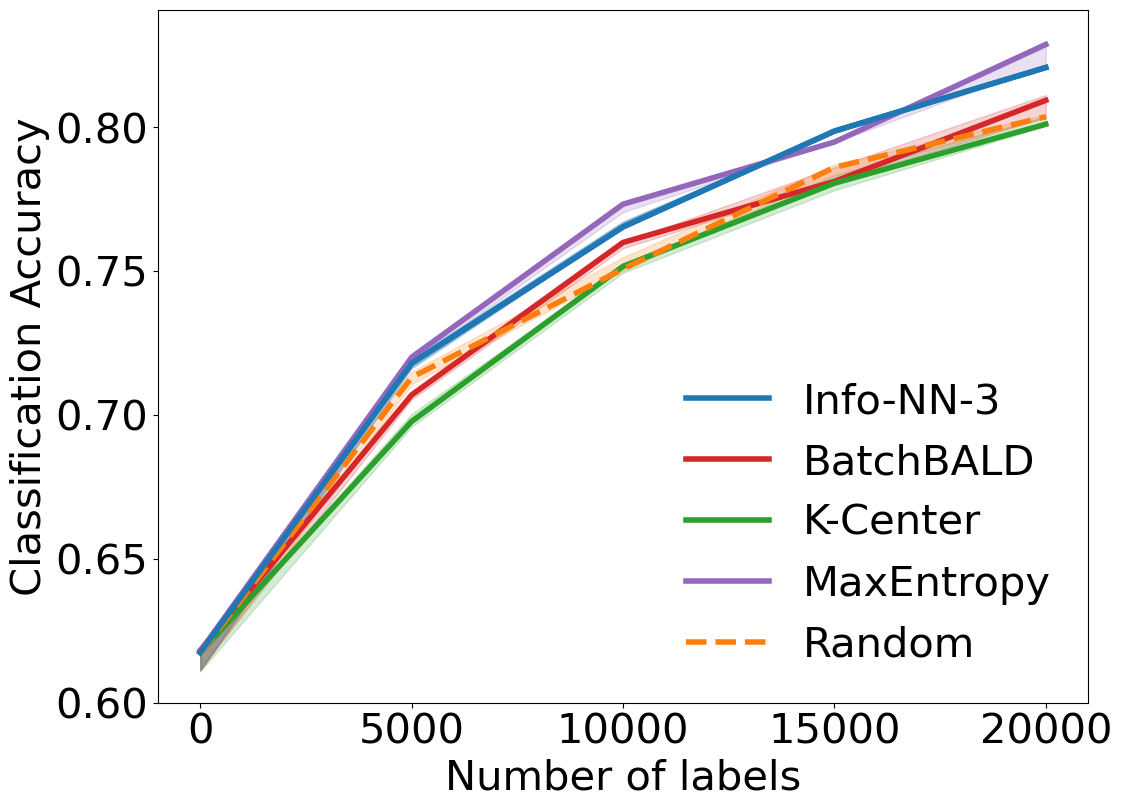}
        \label{fig:cifar10_baselines}
    \end{minipage}
    \hfill
    \begin{minipage}{0.27\linewidth}
        \centering
        \includegraphics[width=\textwidth]{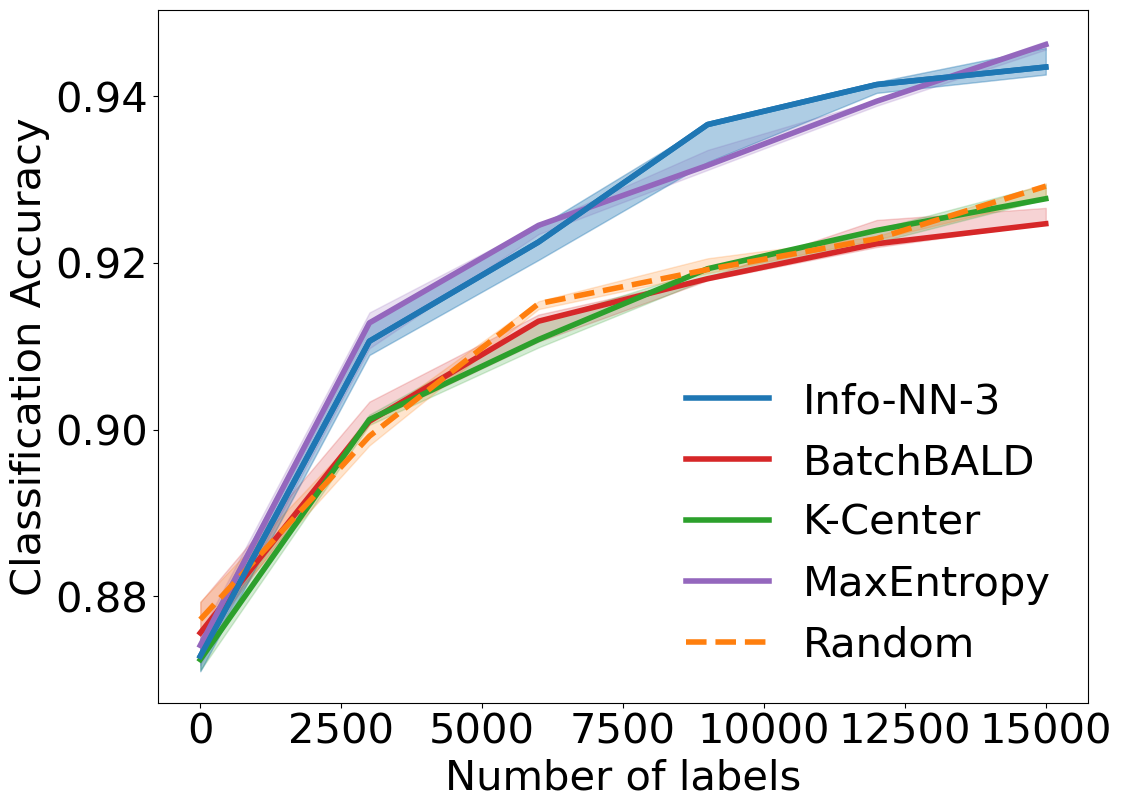}
        \label{fig:svhn_baselines}
    \end{minipage}
    \hfill
    \vskip 0pt
    \vspace{-\baselineskip}
\caption{\small Active classification performance comparison on MNIST (left), CIFAR-10 (center) and SVHN (right) datasets.}
\label{fig:classification_experiments}
\vspace{-\baselineskip}
\end{figure*}

We conduct experiments on the MNIST \cite{lecun1998gradient}, CIFAR-10 \cite{krizhevsky2009learning} and SVHN \cite{netzer2011reading} datasets using CNNs to demonstrate the performance of our active learning method with supervised classification. The experiments on MNIST have an initial balanced labelled set of 30 samples, 3 from every class, chosen at random and an acquisition batch of size 10 is used. For CIFAR-10 and SVHN, we start with initial balanced labelled sets of 5000 and 3000 respectively, and acquire batches of size 5000 and 3000. 
We compare the performance of Info-NN with BatchBALD \cite{kirsch2019batchbald}, $K$-Center \cite{sener2018active}, MaxEntropy, and Random methods. While our method outperforms all the baselines on MNIST, on CIFAR-10 and SVHN, it performs almost on par with MaxEntropy.% and better than the other methods. 

%% file: files/06-conclusion.tex
In this paper, we introduce a generalized similarity based active learning framework for selecting informative queries for both metric learning and classification. In a deep metric learning setting, we demonstrated that our framework is capable of outperforming recently developed methods for selecting batches of triplets on a both per-triplet and per-query basis. For classification, our framework for active label selection resulted in a better performance compared to the baselines. As shown by strong empirical performance, this framework marks the first step in developing a generalized active learning methods capable of performing well in multiple problem areas.

%% file: files/07-supp.tex
%\section{Limitations, societal impacts and datasets}
%\input{files/supp-material/01-checklist_sections}

\section{Plackett-Luce model details}
\input{files/supp-material/01-PL_details}

\section{Computation of mutual information}
\input{files/supp-material/02-MI_comp}

%Metric learning: experimental setup, more results
\section{Metric learning} 
In this section, we provide precise experimental details and highlight additional metric learning experimental results for both DML and non-parametric embedding learning via MDS.
\subsection{Deep metric learning}
\input{files/supp-material/03-1-dml}

\subsection{MDS embedding learning}
\input{files/supp-material/03-2-mds}

%Classification: experimental setup, more results
\section{Classification}

\input{files/supp-material/04-exp_details_classification}

%% file: files/supp-material/01-PL_details.tex
The Plackett-Luce model is derived from an assumption, the Luce's choice axiom \citep{luce1959individual}, also known as the Independence of Irrelevant Alternatives (IIA), which states that the presence of other items in a choice set do not change the relative probabilities of choosing items in the set. This is a reasonable assumption in our setting. This model belongs to a family of discrete choice models which are commonly used to describe situations where a selection is made from a set of options. Such scenarios are encountered widely in the fields of economics \citep{train2009discrete}, for example, to explain the choice made by a company on whether or not to launch a product into the market, in psychology \citep{tversky1981framing} to interpret the choices made by humans in every day situations and, more recently, in computer science \citep{rosenfeld2020predicting} to model choices made by a user in online platforms. 

%% file: files/supp-material/02-MI_comp.tex
We make the following simplifying assumptions to enable efficient computation of the mutual information in practice. We follow a similar approach as the one presented in \cite{canal2019active} and detail it in the context of NN queries.

To derive Info-NN-embedding, we make the following assumptions: 
\begin{enumerate}
    \item The response $Y_n$ is independent of past responses $y^{n-1}$, when conditioned on $\mZ$. 
    \item The oracle's response conditioned on $\mZ$, depends only on $\mZ_{Q_n}$ - embeddings of the items involved in the query and is independent of the embeddings $\mZ_{i \notin Q_n}$.
    \item $\mZ$ is independent of $y^{n-1}$. given the previous estimate of the embedding $\mZ^{n-1}$.
    \item Conditioned on $\mZ^{n-1}$, the $(i, j)^{th}$ entry of $\mZ$, $\mZ_{i,j}$, is distributed normally with mean $\mZ^{n-1}_{i,j}$ and variance $\sigma^2$. We will slightly abuse notation, and write $\mZ \sim \mathcal{N}(\mZ^{n-1}, \sigma^2)$.
\end{enumerate}
% These assumptions are common in the active learning literature, e.g., \citep{canal2020active}.

 We can now re-write $H[Y_n    \: | \:   y^{n-1} ]$ as follows.
 \begin{align}
     H[Y_n    \: | \:   y^{n-1} ] &= H[\underset{\mZ}{\E}(p(Y_n | \mZ, y^{n-1})| y^{n-1})]\\
     &= H[\underset{\mZ}{\E}(p(Y_n | \mZ)| y^{n-1})]\\
     &= H[\underset{\mZ_{Q_n}}{\E}(p(Y_n | \mZ_{Q_n})| y^{n-1})]\\
     &= H[\underset{\mZ_{Q_n}}{\E}(p(y_n | \mZ_{Q_n})| \mZ^{n-1})]\\
     &= H[\underset{\mZ_{Q_n} \sim \mathcal{N}(\mZ_{Q_n}^{n-1}, \sigma^2)}{\E}(p(Y_n | \mZ_{Q_n}))]
 \end{align}
 Following a similar process, we have 
\begin{equation}
    \underset{\mZ}{\E} (H[Y_n    \: | \:   \mZ, y^{n-1} ]) = \underset{\mZ_{Q_n} \sim \mathcal{N}(\mZ_{Q_n}^{n-1}, \sigma^2)}{\E}(H[p(Y_n | \mZ_{Q_n})]).
\end{equation}
We can now utilize Monte Carlo sampling methods for tractable probabilistic inference, as presented in Alg. \ref{alg:info-nn-emb}

For Info-NN-distances, we make the same assumptions as above, except for assumption 4. Instead, we assume that the distances between data points are distributed normally with the mean for each pair set equal to the distance computed from the estimated embedding matrix and variance set to the sample variance of all possible pairwise distances. This assumption enables an efficient method of estimating the posterior distribution over the distances. 

These, along with assumptions on conditional independence of the oracle responses and the distance estimates with respect to the previous responses $y^{n-1}$, enable efficient estimation of the mutual information.   Specifically, the entropies in Eq.~\ref{eq:MI_2} can be computed as follows: 
\begin{align}
H[Y_n    \: | \:   y^{n-1} ] &= H \left[ \underset{\mZ}{\E} \left( p(Y_n    \: | \:   \mZ, y^{n-1} )   \: | \:   y^{n-1}  \right) \right] \notag \\ 
&= H \left[ \underset{D_{Q_n} \sim \mathcal{N}_{Q_n}^{n-1}}{\E} \left( p(Y_n    \: | \:   D_{Q_n}) \right) \right]\label{eq:ent_1}
\end{align}
and
\begin{align}
\underset{\mZ}{\E} (H[Y_n    \: | \:   \mZ, y^{n-1} ]) &= \underset{\mZ}{\E} \left( H \left[ p(Y_n    \: | \:   \mZ, y^{n-1} )   \: | \:   y^{n-1}  \right] \right) \notag \\
&= \underset{D_{Q_n} \sim \mathcal{N}_{Q_n}^{n-1}}{\E} \left( H \left[ p(Y_n    \: | \:   D_{Q_n}) \right] \right)\label{eq:ent_2},
\end{align}
where $D_{Q_n}$ refers to the set of distances between the reference $r_n$ and each of $t_n^c \in T_n$ and $\mathcal{N}_{Q_n}^{n-1}$  represents the assumed normal distribution on $D_{Q_n}$ with the mean and variance determined by the estimate of distances after $n-1$ queries. Due to this normal distribution assumption, the entropy in~\eqref{eq:ent_1} and the expectation in~\eqref{eq:ent_2} are straightforward calculations. The full procedure is shown in Alg.~\ref{alg:info-nn-dist}

\begin{comment}
\begin{algorithm}
\caption{Info-NN-distances}
\label{alg:compute_mi_2} 
\textbf{Input:} Embedding $\mZ$, candidate queries $Q$, num. samples $n_s$, variance $\sigma^2$
\begin{algorithmic}
    \STATE $I \leftarrow \{\}$ (Mutual information values for candidate queries)
    \FOR{$q \in Q$}
    \STATE $r \leftarrow $ first element of $Q$
    \STATE $T \leftarrow Q \backslash \{r\}$
    \STATE $D_q \leftarrow$ distance of every item in $T$ to $r$ in $\mZ$
    \STATE $Y_q \leftarrow$ query response using $D_q$
    \STATE Generate $n_s$ samples of $D_s \sim \mathcal{N}(D_q, \sigma^2)$
    \STATE $I_q \leftarrow H \left[ \underset{D \in D_s}{\E} \left( p(Y_q    \: | \:   D) \right) \right] - \underset{D \in D_s}{\E} \left( H \left[ p(Y_q    \: | \:   D) \right] \right)$
    \STATE $I \leftarrow I \cup I_q$
    \ENDFOR\\
\end{algorithmic}
\textbf{Output:} $I$
\end{algorithm}
\end{comment}

%% file: files/supp-material/03-1-dml.tex
\paragraph{Neural network architectures and learning rates.}
For the DML experiments, we utilize the following network architectures and learning rates for the three datasets. We utilize networks consisting only of fully connected layers with ReLU nonlinearities inserted between all layers.
\begin{itemize}[leftmargin=*]
    \item \textbf{Mahalanobis Metric Dataset: } Fully connected layers of sizes 32, 48, and 10, respectively. Learning rate: $0.0001$
    \item \textbf{Food73 Dataset: } Fully connected layers of sizes 12, 12, and 12, respectively. Learning rate: $0.0005$
    \item \textbf{Graduate Admissions Dataset: } Fully connected layers of sizes 16, 12, and 10, respectively. Learning rate: $0.0001$
\end{itemize}
We utilize the same learning rate for re-training models across all methods (random, Info-NN, Batch-Euclidean/Centroid).

\paragraph{Experiment parameters.}
In all experiments, we utilized a value of $\mu = 0.00001$ for the probability model and utilized $20$ initialization triplets. Batch sizes of $10$ (synthetic), $30$ (food and graduate admissions) are used. Furthermore, for Info-NN experiments, we utilize the following values for hyperparameters $\sigma^2$ (distance distribution variance), $n_s$ (number of samples used to compute mutual information), $B$ and $B^\prime$ (number of top most informative queries selected per batch):
\begin{itemize}[leftmargin=*]
    \item \textbf{Synthetic Mahalanobis Metric Dataset: } $\sigma^2 = 1$, $n_s = 100$, $B^\prime = 10 = B$
    \item \textbf{Food73 Dataset: } $\sigma^2 = 6.5$, $n_s = 1,000$, $B^\prime = 5$
    \item \textbf{Graduate Admissions Datset: } $\sigma^2 = 10$, $n_s = 1,330 ~(= 10N)$, $B^\prime = 5$
\end{itemize}
As reported in the main paper, we used batch sizes of $10, 30,$ and $30$ for the Mahalanobis, food, and admissions datasets respectively. These batch sizes are the sizes of the NN queries collected. For any method using triplets, the batch size is doubled, resulting in batch sizes of $20, 60,$ and $60$, respectively. This is done so we can compare both on a per-query and per-triplet basis. To set such parameters, a coarse grid search was performed to find the best performing parameters. 

We compared our method against two baselines found in \cite{kumari2020batch}. These baselines follow the same general approach of weighting informativeness (measured using entropy) and diversity (measured using various metrics such as the Euclidean distance of all permutations of the triplet or the centroid of the three points selected in the triplet) for an \textit{overcomplete} batch size. We utilize an overcompleteness factor of $3$, which indicates that for a batch of $B$ triplets, the $3B$ most informative triplets are identified. The informativeness of the $3B$ triplets are then weighted by the informativeness, and the top $B$ triplets are then presented to the oracle. From studies performed by \cite{kumari2020batch}, anything above a factor of $2$ exhibits roughly the same performance.

\begin{figure}[t]
\captionsetup[sub]{justification=centering}
\begin{subfigure}{\textwidth}
    \centering
    \includegraphics[scale=0.25]{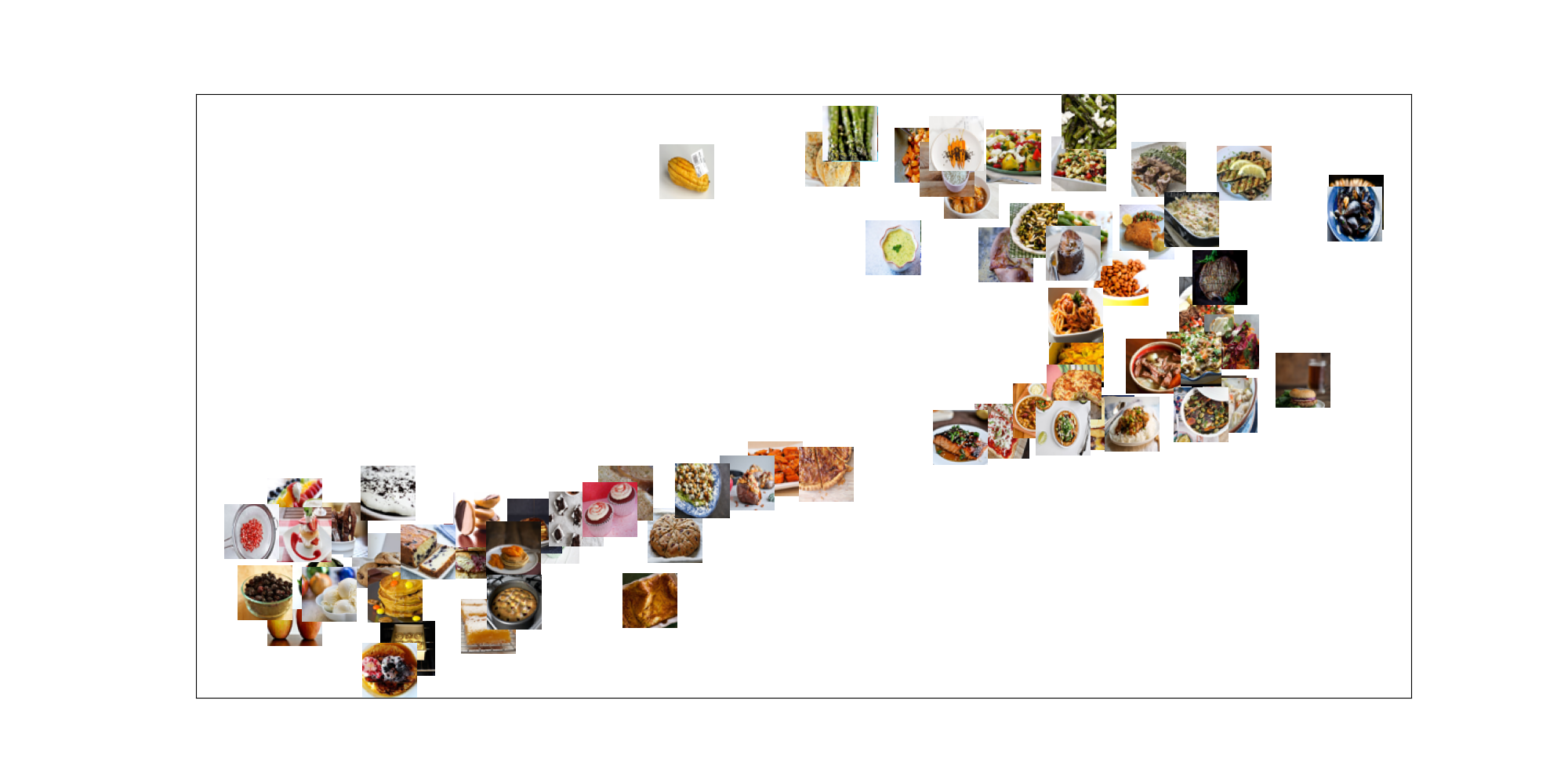}
\end{subfigure}
\vskip 0pt
\vspace{-1\baselineskip}
\begin{subfigure}{\textwidth}
    \centering
    \includegraphics[scale=0.25]{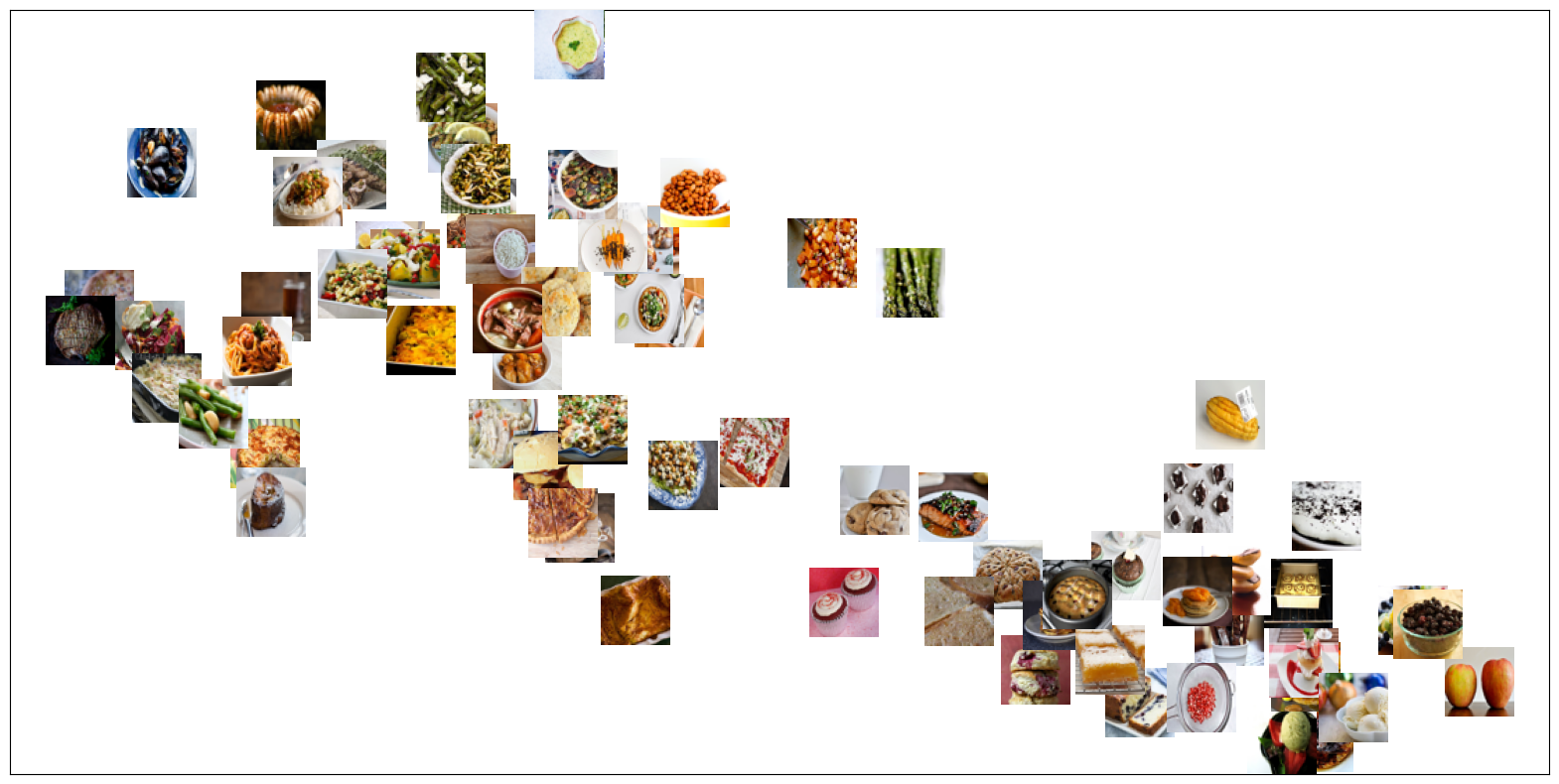}
\end{subfigure}
\caption{Visualization of food embedding learned using queries selected with Batch-Centroid (top) and Batch-Euclidean (bottom) generated using t-SNE \cite{maaten2008visualizing}.}
\label{fig:food_viz_batch}
\end{figure}

\paragraph{Additional embedding visualizations.}
Models used to generate all embedding visualizations, including those shown in the main paper, used the same number of triplets. We present an additional visualization of the Food73 dataset embedding learned with the Batch-Centroid and Batch-Euclidean methods in Fig. \ref{fig:food_viz_batch}. In comparison to the embedding learned with Info-NN (Fig. 3 in main paper), the embedding learned with Batch-Centroid after the same number of triplets does a poorer job of grouping together vegetables, unlike the Info-NN embedding. 

\begin{figure}
    \centering
    \includegraphics[width=0.4\textwidth]{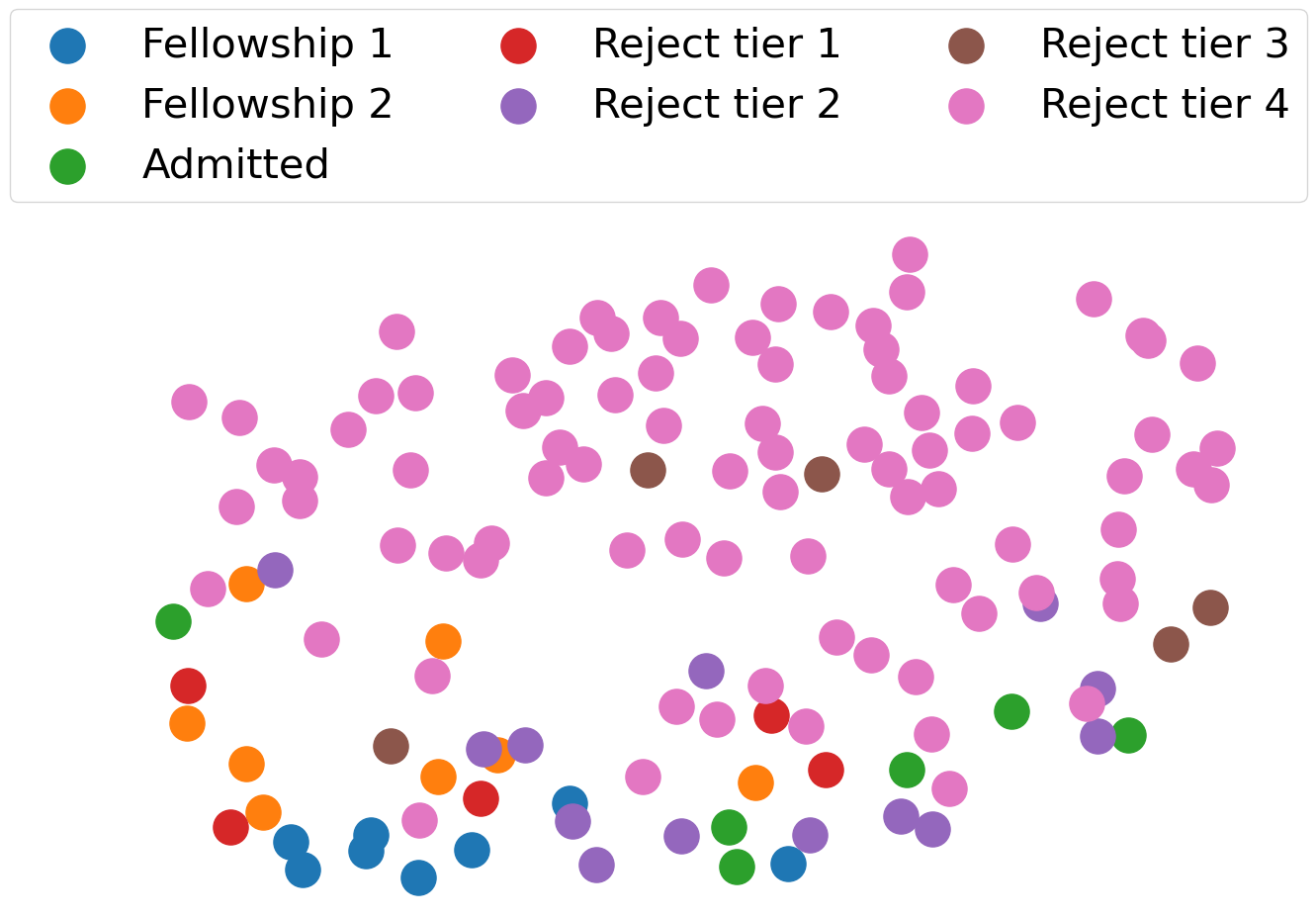}
    \caption{Visualization of admissions embedding learned using queries selected with Batch-Euclidean generated using t-SNE \cite{maaten2008visualizing}.}
    \label{fig:adm_vis_euclid}
\end{figure}

We also present a visualization of the embedding learned via Batch-Euclidean on the Graduate Admissions dataset in Fig. \ref{fig:adm_vis_euclid}. Comparing embeddings learned with Info-NN and Batch-Centroid (Fig. 5 in main paper) and Batch-Euclidean, it is clear that Info-NN selects queries that more closely group highly ranked candidates together. However, none of the methods visualized are able to completely cluster candidate tiers distinctly; for all three methods, admitted students (fellowship and non-fellowship) are intermingled with candidates in the first and second rejection tiers. %This supports the idea discussed in Sec. \ref{sec:society} that in many scenarios, raw data features are unable to tell the full story. We did not consider parts of the application, such as Personal Statement or Statement of Purpose, which allow applicants to put in explanatory personal details. 

\begin{figure}[t]
\captionsetup[sub]{justification=centering}
\hspace*{\fill}
\begin{subfigure}{0.31\textwidth}
    \includegraphics[width=\textwidth]{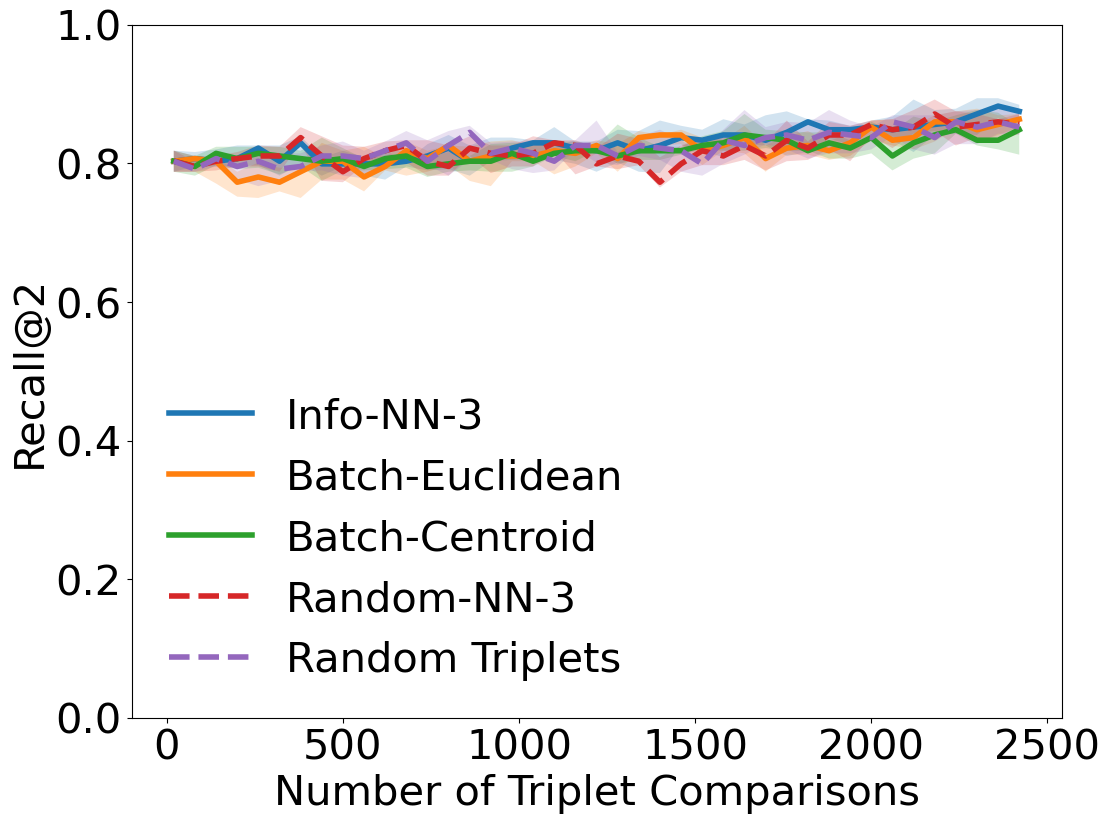}
    \label{fig:dml_adm_recall2}
\end{subfigure}
\hspace*{\fill} % separation between the subfigures
\begin{subfigure}{0.31\textwidth}
    \includegraphics[width=\textwidth]{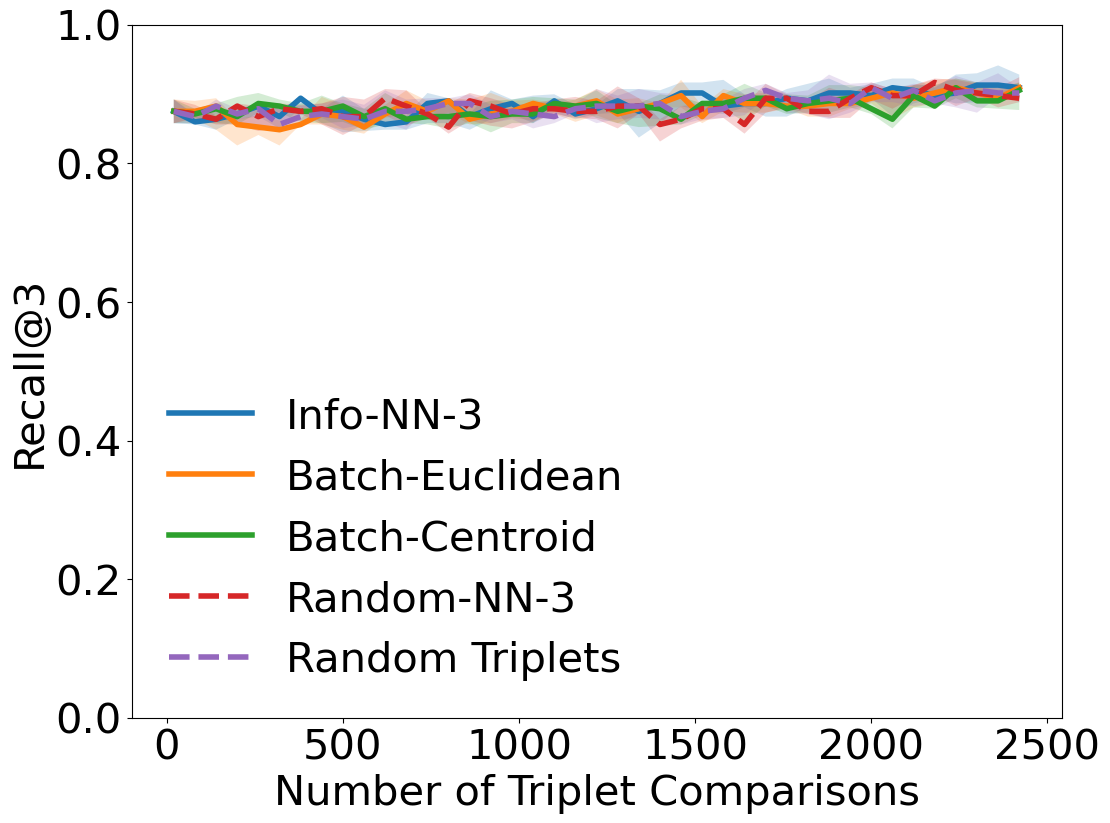}
    \label{fig:dml_adm_recall3}
\end{subfigure}
\hspace*{\fill}
\vskip 0pt
 \vspace{-1\baselineskip}
\hspace*{\fill}
\vspace{-1\baselineskip}
\begin{subfigure}{0.31\textwidth}
    \includegraphics[width=\textwidth]{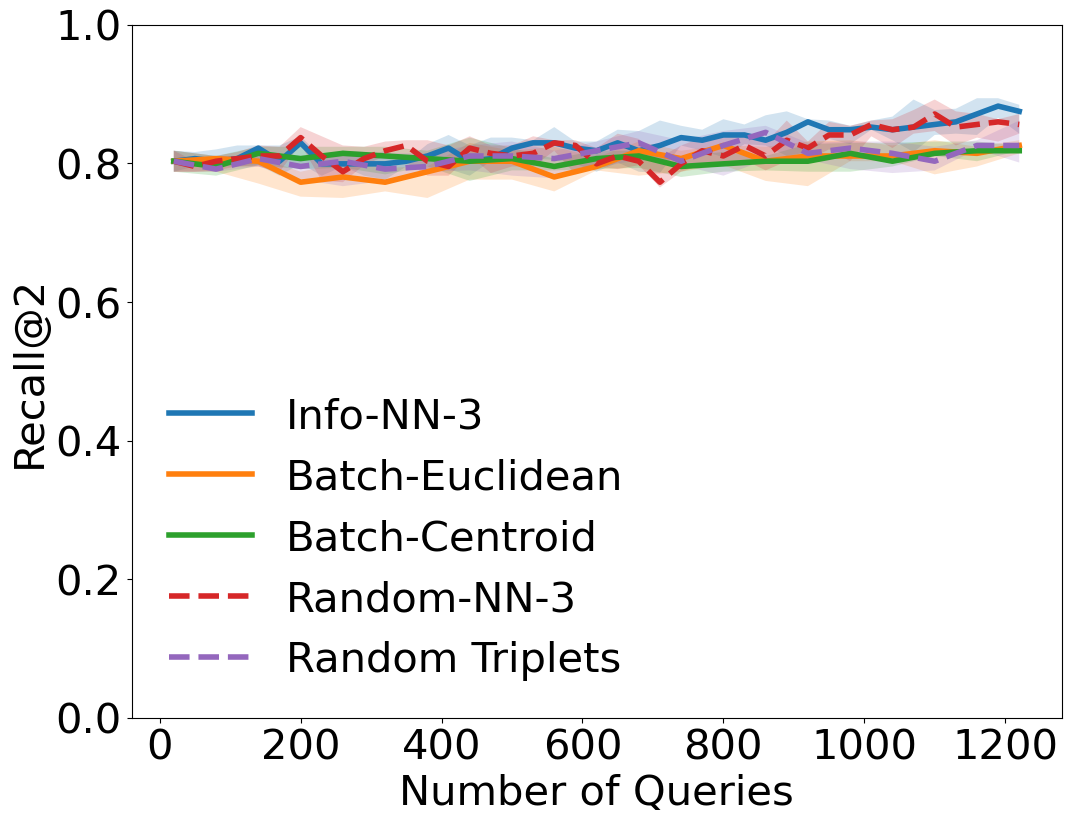}
    \label{fig:dml_adm_recall2_queries}
\end{subfigure}
\hspace*{\fill}
\begin{subfigure}{0.31\textwidth}
    \includegraphics[width=\textwidth]{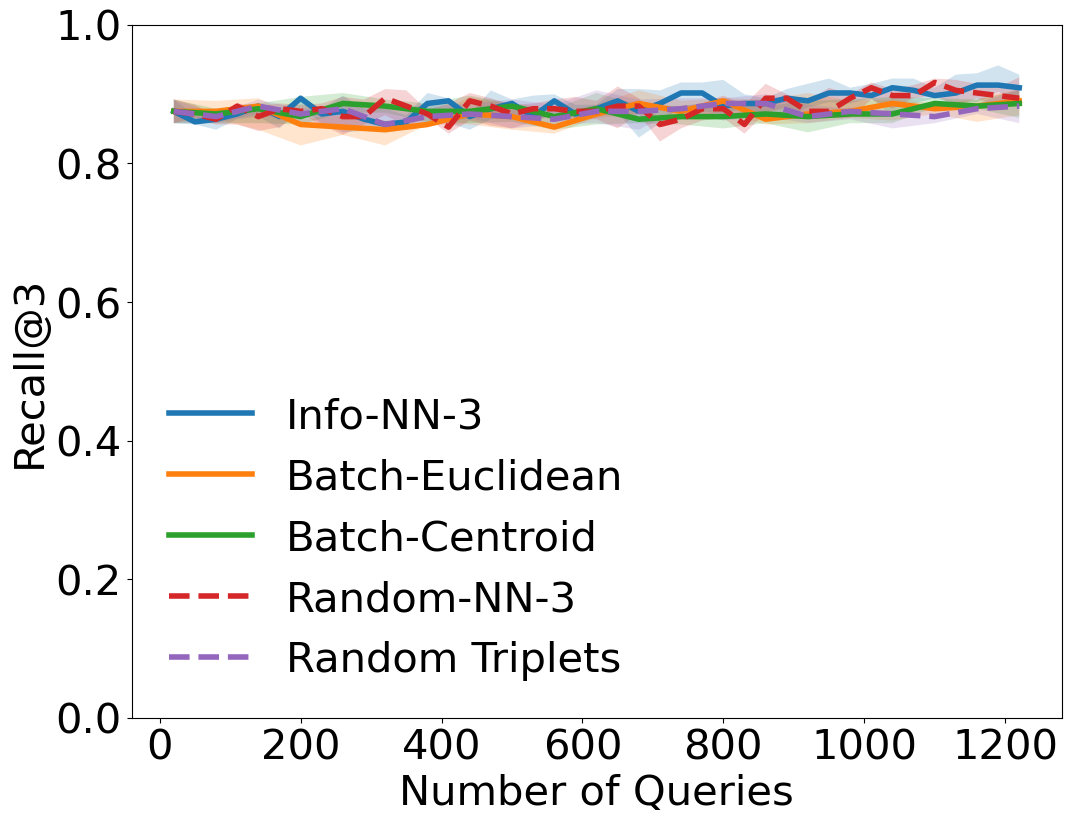}
    \label{fig:dml_adm_recall3_queries}
\end{subfigure}
\hspace*{\fill}
%\vspace{-1\baselineskip}
\caption{\small Per-triplet (top) and per-query (bottom) comparison for Info-NN against other methods. Recall@$2$ (left) and Recall@$3$ (right).} 
\label{fig:dml_recall23}
\end{figure}

\begin{figure}[t]
\captionsetup[sub]{justification=centering}
\hspace*{\fill}
\begin{subfigure}{0.31\textwidth}
    \includegraphics[width=\textwidth]{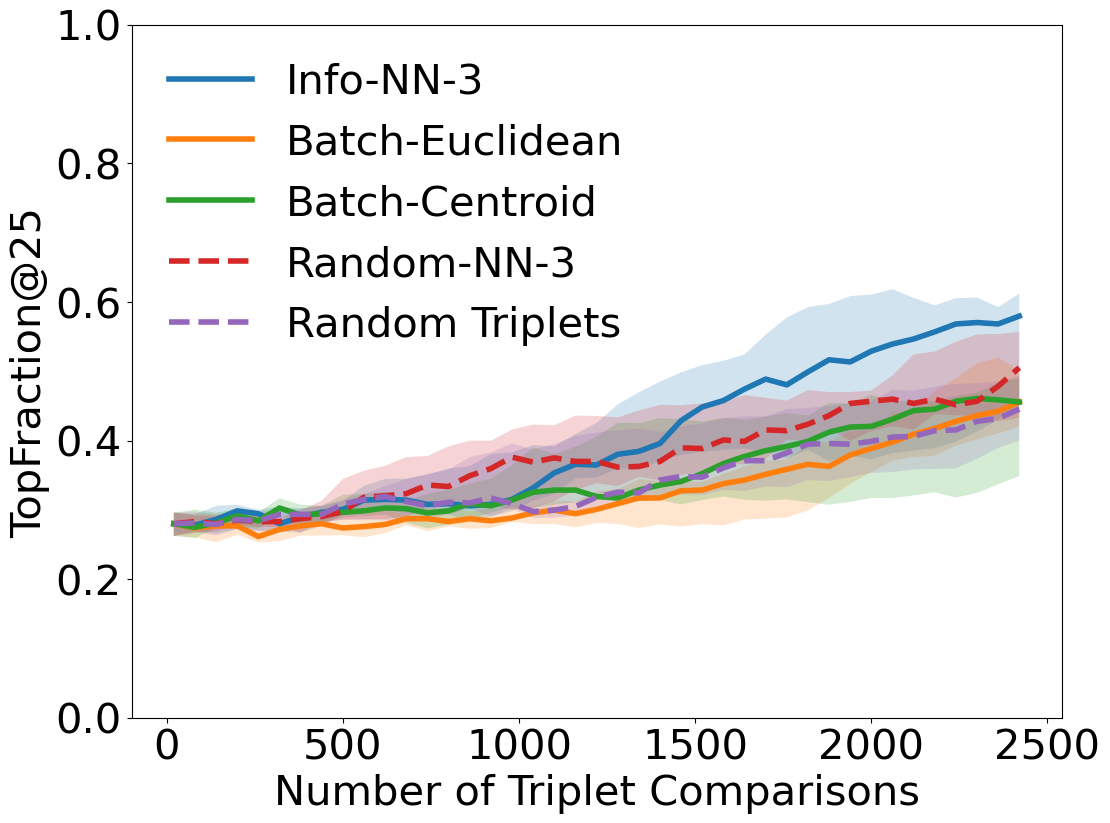}
    \label{fig:dml_adm_frac2}
\end{subfigure}
\hspace*{\fill} % separation between the subfigures
\begin{subfigure}{0.31\textwidth}
    \includegraphics[width=\textwidth]{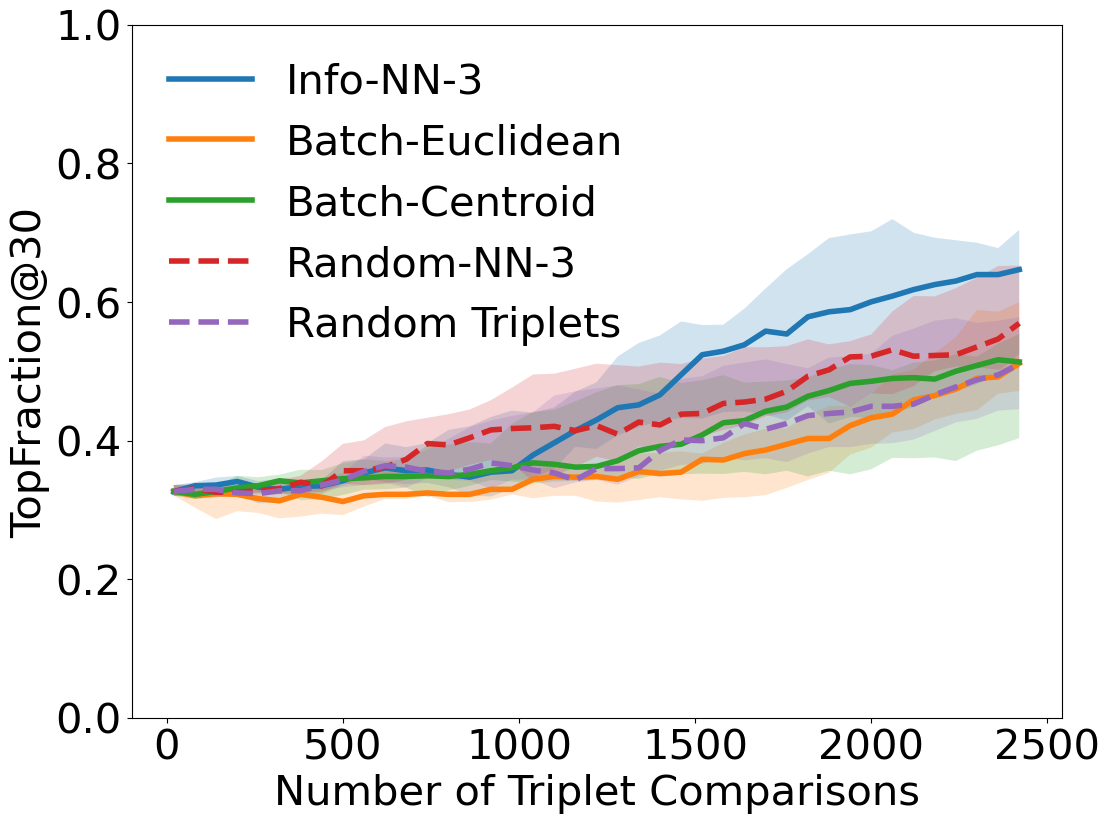}
    \label{fig:dml_adm_frac3}
\end{subfigure}
\hspace*{\fill}
\vskip 0pt
 \vspace{-1\baselineskip}
\hspace*{\fill}
\vspace{-1\baselineskip}
\begin{subfigure}{0.31\textwidth}
    \includegraphics[width=\textwidth]{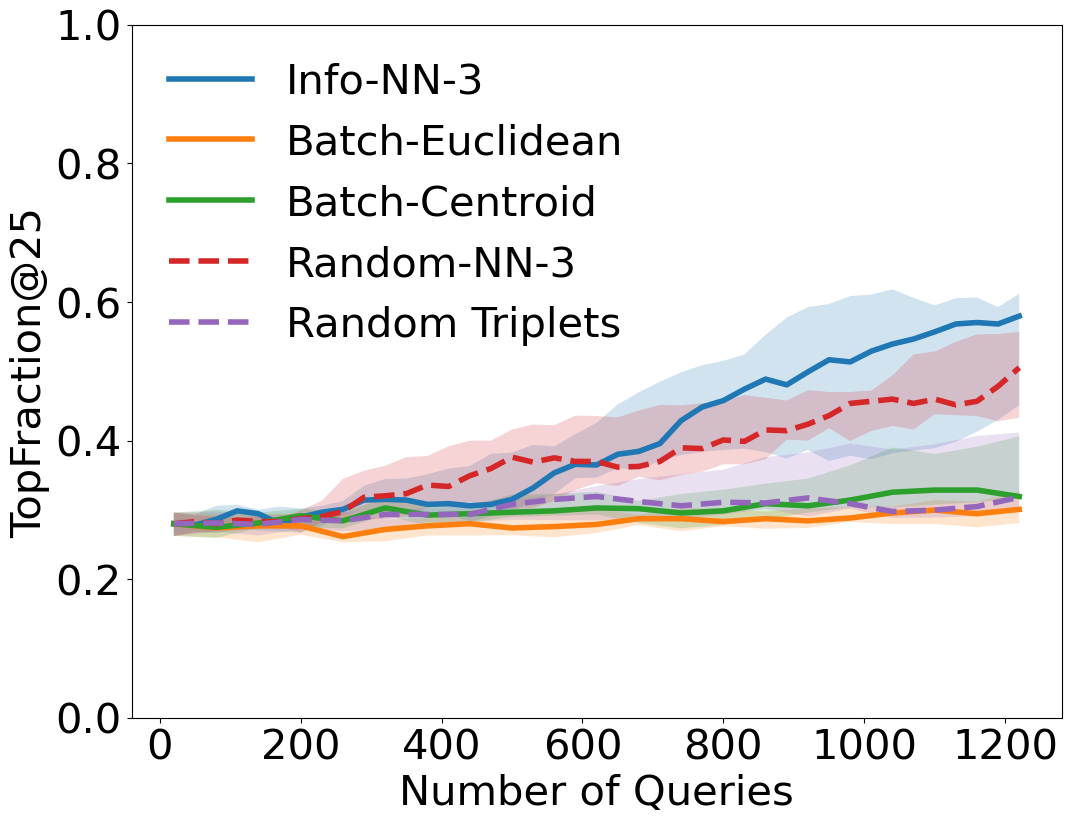}
    \label{fig:dml_adm_frac2_queries}
\end{subfigure}
\hspace*{\fill}
\begin{subfigure}{0.31\textwidth}
    \includegraphics[width=\textwidth]{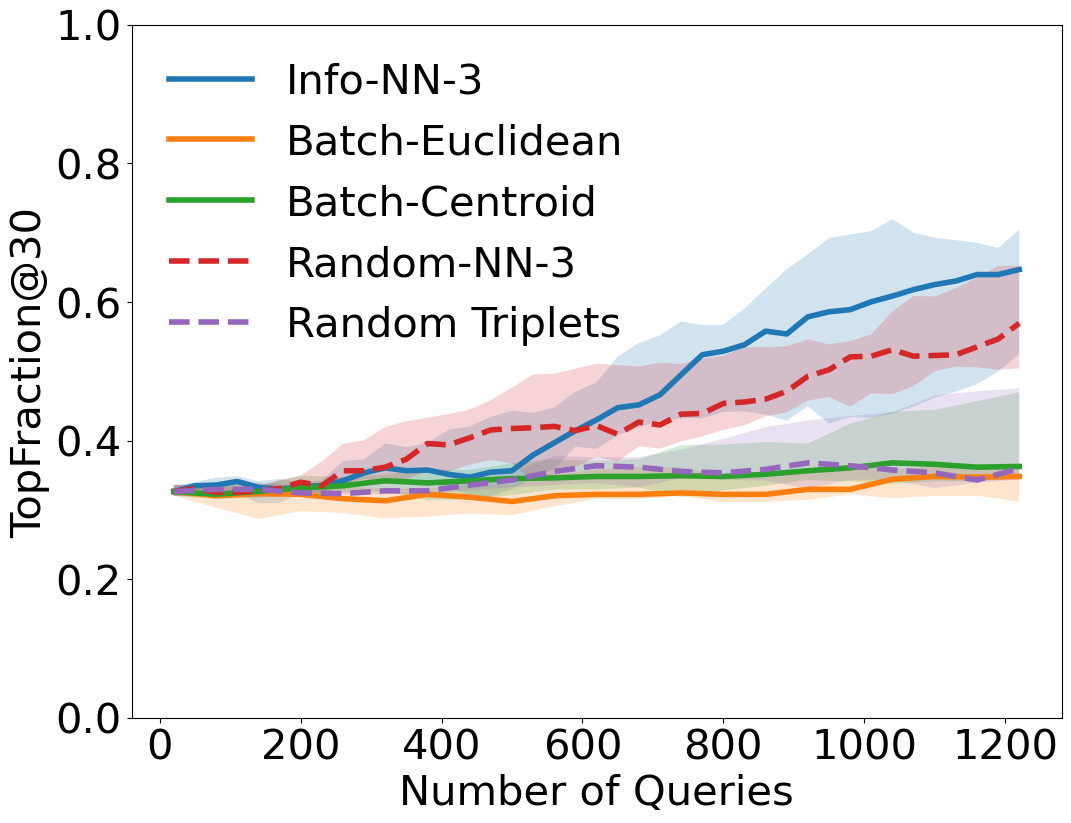}
    \label{fig:dml_adm_frac3_queries}
\end{subfigure}
\hspace*{\fill}
%\vspace{-1\baselineskip}
\caption{\small Per-triplet (top) and per-query (bottom) comparison for Info-NN against other methods. TopFraction@$25$ (left) and TopFraction@$30$ (right).} \vspace{-5mm}
\label{fig:dml_frac23}
\end{figure}

\paragraph{Additional results on Graduate Admissions dataset.}
Results for additional values of $K$ for Recall@$K$ and TopFraction@$K$ are presented in Fig. \ref{fig:dml_recall23} and Fig. \ref{fig:dml_frac23}, respectively. On both a per-triplet and per-query basis, Info-NN is performs the best for all values of $K$. We note that for Recall@$K$ for larger values of $K$, all methods perform roughly the same and perform well. This is because the dataset contains a large number of tier 4 rejections, which every method is able to successfully group together, inflating the Recall@$K$ value. Thus, we believe that the TopFraction@$K$ results do a better job of illustrating how the method does in selecting queries that group admitted or more highly ranked candidates together. 

%% file: files/supp-material/03-2-mds.tex
We perform a set of experiments which utilize MDS to learn representations of the items. In particular, we use this opportunity to compare the performance of NN queries against a more complex ranking query \cite{canal2020active}. When comparing against ranking queries, it is important to note that \textbf{we expect both actively selected and randomly selected ranking queries to outperform a nearest neighbor query of the same size on a per-query basis}, as there is a discrepancy in the amount of information each query contains. All experiments were performed on a 2019 MacBook Pro, 2.6 GHz 6-Core Intel i7, 16 GB RAM.

\paragraph{Data generation.}
In each simulation, the ground truth embedding consists of points drawn independently from a multivariate Normal distribution with mean $\vzero$ and covariance matrix $\mtx{I}$. We utilize a deterministic oracle, which orders the items based on their true distances from the selected reference object and generate a new initialization embedding with entries drawn uniformly at random from $[0,1]$ for every trial.

\paragraph{Experiment parameters.} 
For both the Info-NN vs. Random-NN and Info-NN vs. Ranking experiments, we utilize a diminishing $\mu$ parameter. For each active learning iteration $k \in \{1, \ldots, K\}$, we set $\mu = D_{\text{max}}(0.99)^k$, where $D_{\text{max}}$ is the maximum pairwise distance in the current estimate of the embedding. As presented in \citep{tamuz2011adaptively}, the $\mu$ parameter can be thought of as a margin. With a diminishing $\mu$, we are enforcing a stricter margin in the earlier stages of learning, when our estimate of the embedding is poor. As the number active learning cycles increases, our estimate of the embedding should improve, thus lessening the need for a larger margin. Multiple other options for $\mu$ were considered, such as setting $\mu$ to a constant or the maximum of all pairwise distances, but we found that the diminishing $\mu$ worked well for the MDS synthetic embedding learning experiments.

We utilized step size of $\alpha = 0.5$ for probabilistic MDS. This parameter was not finely tuned. We observed similar performance as long as $\alpha$ is reasonably small ($\alpha < 1$). 

\paragraph{Probabilistic multidimensional scaling.}
To fit an embedding using nearest neighbor or ranking queries, we first decompose the query response into a set of paired comparisons and store these paired comparisons in $\mathcal{S}$. A nearest neighbor query of size $C$ as $C-1$ paired comparisons and similarly, ranking query of size $C$ can be decomposed into $\frac{C(C-1)}2$ paired comparisons. Thus, the active embedding technique framework is general enough to accommodate both query types. We then utilize a version of the probabilistic multidimensional scaling (MDS) approach presented in \citep{tamuz2011adaptively}. Starting with some input embedding $\mZ$, we perform a fixed number of gradient descent iterations with a fixed step size $\alpha$ (not necessarily to convergence) on the empirical log-loss   
\[
\ell_{\mathcal{S}}(\mZ) = \frac1{|\mathcal{S}|}\sum\limits_{i=1}^{|\mathcal{S}|} \log\frac{1}{P_{Q_i}},
\] where for $Q_i = r_i \cup \{t_i^1, t_i^2\}\in \mathcal{S}$
\[
P_{Q_i}(Y_i = t_i^1) = \frac{(D_{r_i,t_i^1}^2 + \mu)^{-1}}{(D_{r_i,t_i^1}^2 + \mu)^{-1} + (D_{r_i,t_i^2}^2 + \mu)^{-1}}.
\] 
That is, we perform updates of the form $\mZ = \mZ - \alpha\nabla \ell_{\mathcal{S}}(\mZ)$. 

Our active embedding strategy, utilizing probabilistic MDS, is as follows: Starting with an initial embedding $\mZ_0$, we initialize our algorithm by running probabilistic MDS on $\mZ_0$ with $K_0$ randomly drawn queries to obtain $\mZ_1$.
At each iteration $k > 0$, we alternate between the following: 
\vspace{-1mm}
\begin{enumerate}
    \item Fix each column in $\mZ_{k}$ as the reference data point, run Info-NN to find the query that maximizes mutual information with respect to the reference, and choose the query with the maximum mutual information over all $N$ reference data points.
    \item Solicit a response from the oracle for the chosen query, append the paired comparison decomposition to $\mathcal{S}$, and apply probabilistic MDS to $\mZ_{k}$ with the updated $\mathcal{S}$ to obtain $\mZ_{k+1}$. 
\end{enumerate}
\vspace{-1mm}
The full procedure can be found in Alg. \ref{alg:active_emb}

\begin{algorithm}[t]
\caption{Info-NN-C: Active Embedding Technique}
\label{alg:active_emb}
\begin{algorithmic}
\REQUIRE Embedding $\mZ_{\text{init}} \in \R^{D \times N}$, query length $C$, number of active learning cycles $K$, burn-in period $K_0$, number of samples $n_s$, number of MDS iterations $K_{\text{MDS}}$, MDS step size $\alpha$
\STATE $\mathcal{S} \leftarrow \{\}$ 
\FOR{$k = 1,\ldots,K_0$}
\STATE $Q_k \leftarrow $ query of size $C$ drawn uniformly at random
\STATE $y_k \leftarrow $ oracle response to $Q_k$
\STATE $\mathcal{S} \leftarrow \mathcal{S} ~\cup (y_k, Q_k)$ 
\ENDFOR
\STATE $\mZ_0 \leftarrow \text{probabilisticMDS}(\mZ_{\text{init}}, \mathcal{S}, K_{\text{MDS}}, \alpha)$
\FOR{$k = 1,\ldots,K$}
\STATE $(I, Q) \leftarrow \{\}$ (Store highest MI value and corresponding query for all references)
\FOR{$j = 1,\ldots,N$}
    \STATE $Q_j \leftarrow$ Set of all queries of size $C$ for which to compute MI with $j$ as reference item
    \STATE $I_j \leftarrow \text{Info-NN-distances}(\mZ, Q_j, n_s)$ (Compute MI for each query)
    \STATE $(I, Q) \leftarrow (I, Q) \cup (\max I_j, \arg\max I_j)$ (Store query in $Q_j$ with highest MI)
\ENDFOR
\STATE $Q_{j^{\star}} \leftarrow $ Query in $(I, Q)$ with highest corresponding value in $I$
\STATE $y_{j^{\star}} \leftarrow $ Oracle response to $Q_{j^{\star}}$
\STATE $\mathcal{S} \leftarrow \mathcal{S} ~\cup (y_{j^{\star}}, Q_{j^{\star}})$
\STATE $\mZ_k \leftarrow \text{probabilisticMDS}(\mZ_{k-1}, \mathcal{S}, K_{\text{MDS}}, \alpha)$
\ENDFOR\\
%\algorithmicoutput ~$\mZ_k$
\end{algorithmic}
\end{algorithm}

\paragraph{Evaluation metrics.}
To quantify the performance of our approach, we examine how well our recovered embedding preserves the rank ordering of the items. To do so, we use the Kendall's Tau rank correlation coefficient \cite{kendall1938new}. To capture the holistic quality of the learned embedding, we set each object as the reference object, rank all other items based on distance to the reference object, and compute the Kendall's Tau between that item and the ranking induced by the ground-truth embedding with the same reference object. We then define the \emph{aggregate Kendall's Tau} as the mean of all of these Kendall's Tau coefficients. In our simulations we consider multiple trials and we report the median aggregate Kendall's Tau and the 25\% and 75\% quantiles. 

For the following experiments, \emph{Info-Ranking-C} means the active selection method in \cite{canal2020active} was used to select ranking queries each with a set $T_i$ of size $C$.

\paragraph{Info-NN vs. Random-NN.} 
In the first simulation, we quantify the improvement in using the adaptive algorithm over randomly selected nearest neighbor queries. In particular, we fix $N = 20$, $D = 2$ or $D = 5$, use $K_0 = 20$ initial random queries, and examine the performance for queries of sizes $K = 2, 3, 4$, and $5$. % For a fixed query size, we perform 20 trials and report the median and 25\% and 75\% quantiles of the aggregate Kendall's Tau. For each trial, we use a new embedding. 

\begin{figure}[t]
\captionsetup[sub]{justification=centering}
\begin{subfigure}{0.43\textwidth}
    \includegraphics[width=\textwidth]{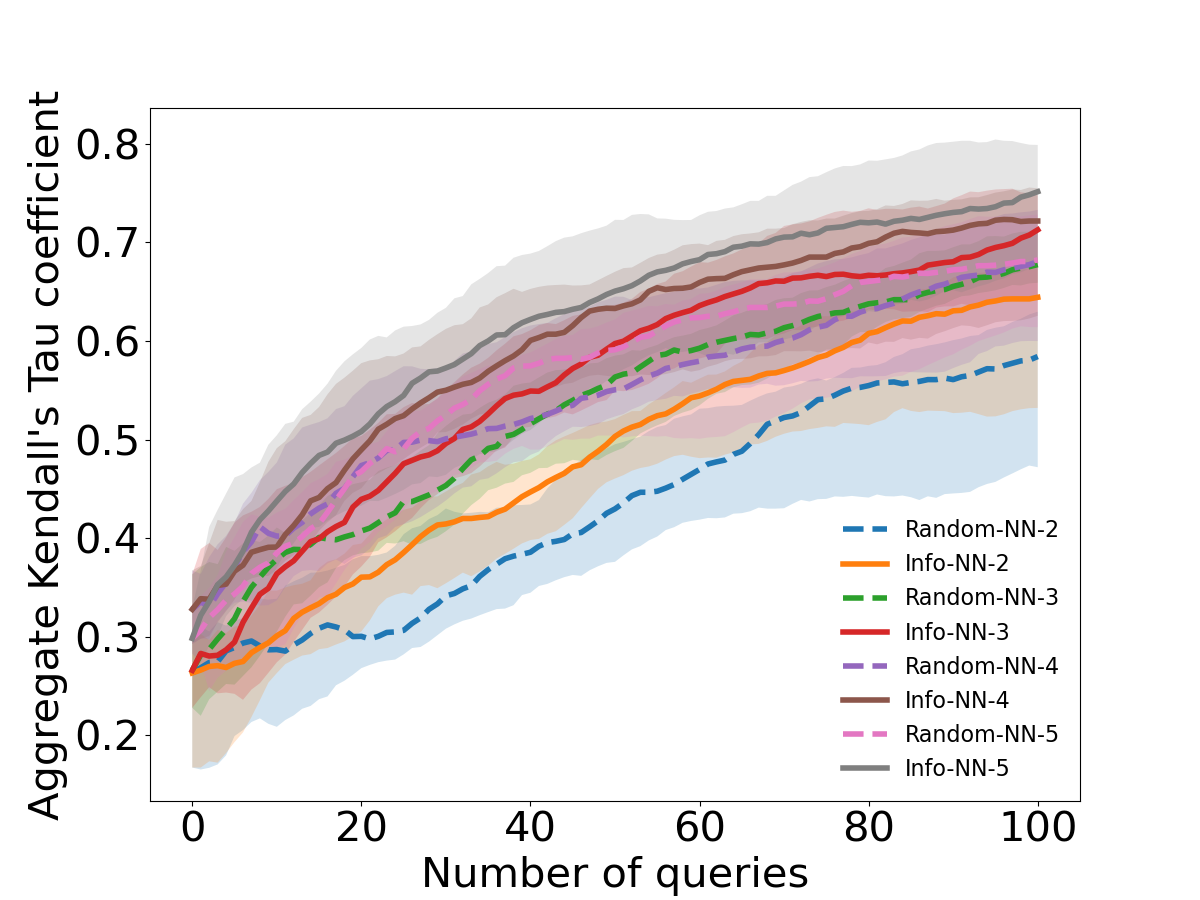}
    \vspace{-1\baselineskip}
\end{subfigure}
\hspace*{\fill} % separation between the subfigures
\begin{subfigure}{0.43\textwidth}
    \includegraphics[width=\textwidth]{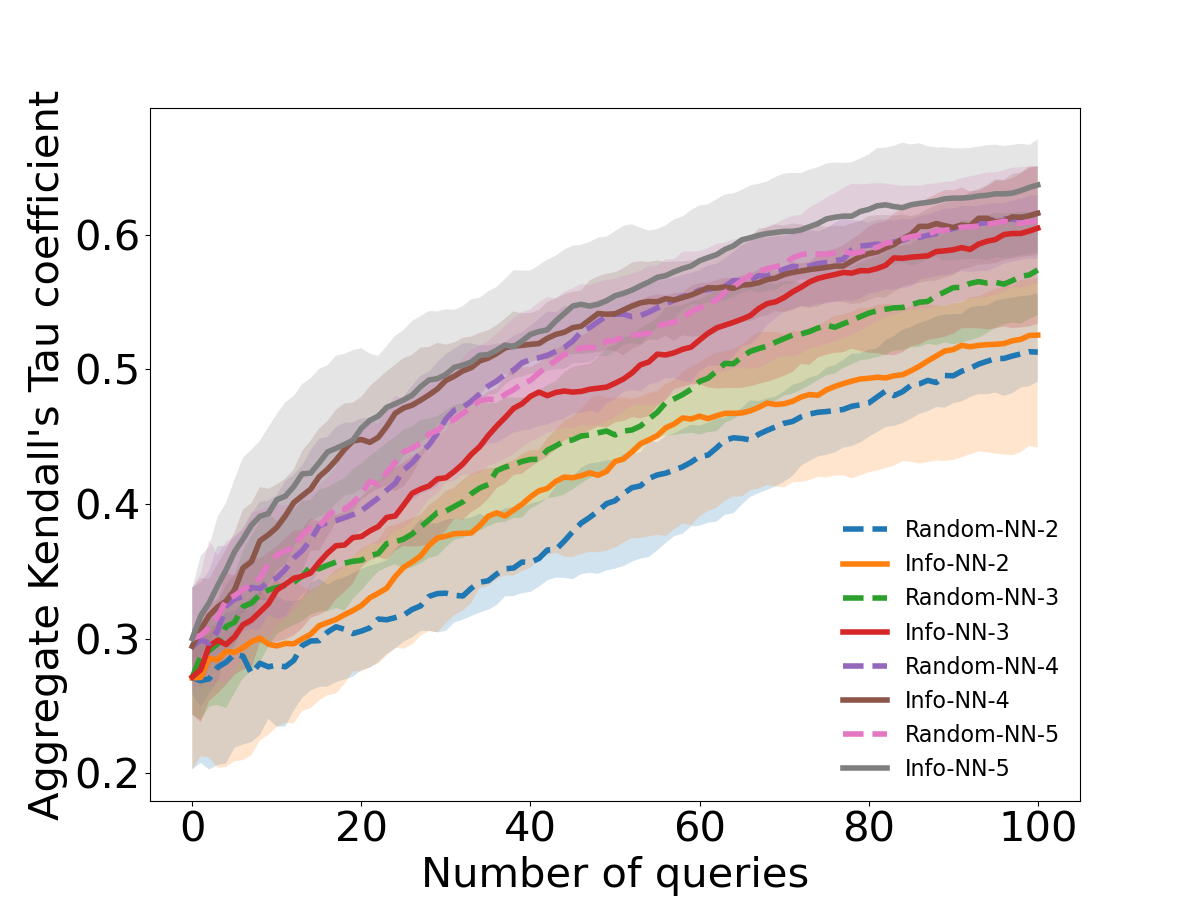}
    \vspace{-1\baselineskip}
\end{subfigure}
\hspace*{\fill}
\caption{\small Comparison of actively selected nearest neighbor queries and randomly selected nearest neighbor queries for $D = 2$ (left) and $D = 5$ (right). Info-NN outperforms randomly selected queries in all cases, even outperforming randomly selected queries of larger size in some cases. Gradient step parameters: $500$ iterations, step size = $0.5$.} \vspace{-5mm}
\label{fig:top1randadapt}
\end{figure}

As shown in Fig.~\ref{fig:top1randadapt}, for all query sizes the learned embedding is significantly better when queries are selected actively rather than at random. Notably, Info-NN-3 queries exceed the performance of randomly selected size 4 and 5 queries despite being smaller. Randomly selected nearest neighbor queries of sizes 3, 4, and 5 all performed similarly, indicating that randomly selected queries contain redundant information that cannot be overcome solely by increasing the query size. %This matches our intuition about \eqref{eq:MI_2} in that our algorithm avoids selecting redundant queries. 

\begin{figure}[t]
 %   \vspace{.3in}
    \centering
    \includegraphics[width=0.43\textwidth]{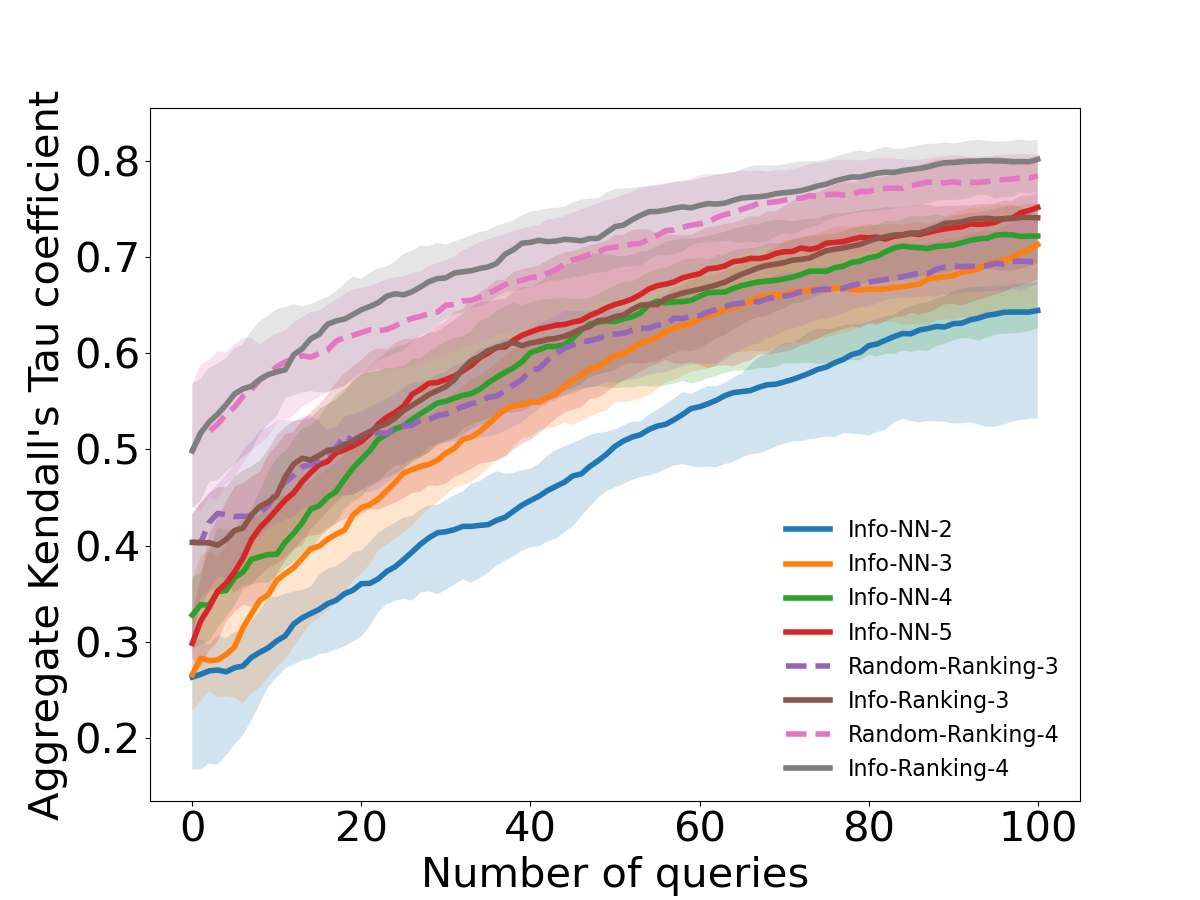}
    %\vspace{.3in}
    \caption{Comparison of actively selected nearest neighbor queries and actively selected and randomly selected ranking queries. Info-NN performs is competitive with a randomly selected ranking query of the same size. Gradient step parameters: $500$ iterations, step size = $0.5$.}
    \label{fig:top1ranking}
\end{figure}

\paragraph{Info-NN vs. Ranking.}
In the second simulation, we compare the performance of actively selected nearest neighbor queries against ranking queries \cite{canal2020active}. We observe that nearest neighbor queries perform competitively to ranking queries, as illustrated in Fig.~\ref{fig:top1ranking}. Again, we fix $N=20$, $D = 2$, utilize $K_0 = 20$ initial random queries, and examine the performance of Info-NN queries of sizes 3, 4, and 5 and ranking queries of sizes 3 and 4. % We report the median and 25\% and 75\% quantiles of the aggregate Kendall's Tau over 20 trials in Fig.~\ref{fig:top1ranking}. Again, we generate a new  embedding for every trial.

We observe that the nearest neighbor query exhibits similar performance to randomly selected ranking queries, despite the ranking queries containing twice as many paired comparisons as a nearest neighbor query. Info-NN-3 queries are able to match randomly selected ranking queries of the same size, while Info-NN-4 queries exceed the performance of randomly selected ranking queries of size 3, while almost matching the performance of actively selected size 3 ranking queries. Employing Info-NN can nearly compensate for the difference in information between nearest neighbor and ranking queries, highlighting an advantage in the trade-off between complexity and ``information density'' (the number of triplets contained in one query).

\subsection{Active selection computational comparison}
While our mutual information computation strategy is similar, utilizing NN queries results in computational advantages when compared to the ranking query used in \cite{canal2020active}. To compare the time discrepancy between computing mutual information for ranking and nearest neighbor queries, we perform $10$ iterations of our embedding technique, and record the amount of time it takes to compute the mutual information for each object as the reference object. We then report the average and standard deviation of the times taken. We use the same parameters for each active learning algorithm, such as number of queries to consider and number of distance samples generated. In Table \ref{tab:rank_nn_timing}, we report the average amount of time it takes to compute the mutual information for a given reference object for differently sized queries in actively selecting nearest neighbor queries using Alg.~\ref{alg:active_emb} and the method presented in \cite{canal2020active}. The drastic discrepancy in timing between the two methods is due primarily to the fact that the nearest neighbor mutual information computation does not require computation for all possible permutations of the set of $K$ items, whereas the ranking query does.

\begin{table}[t]
    \centering
    \caption{Timing results, in seconds, for computing mutual information for nearest neighbor and ranking queries. Experiments performed on 2019 MacBook Pro, 2.6 GHz 6-Core Intel i7, 16 GB RAM.} 
    \vspace{2mm}
    \label{tab:rank_nn_timing}
    \resizebox{0.5\textwidth}{!}{
    \begin{tabular}{l c c c}
        \toprule
        & $K = 2$ & $K = 3$ & $K = 4$\\
        \midrule
        NN & $0.0265 \pm 0.0036$ & $0.1509 \pm 0.0044$ & $0.6634 \pm 0.0812$ \\
        Ranking & $ 0.6605\pm0.0583 $ & $ 8.5394\pm0.3400 $ & $175.0046\pm93.2602 $\\
        \bottomrule
    \end{tabular}}
\end{table}

%% file: files/supp-material/04-exp_details_classification.tex
\subsection{Algorithms}

A description of the active classification framework and the complete Info-NN query strategy utilized to select samples for labelling, is below.

\begin{algorithm}
\caption{Active Learning for Classification}
\label{alg:al_classification}
\begin{algorithmic} 
\REQUIRE Dataset $\mathcal{X} = \{\vx_i\}_{i=1}^N$, batch size $b$, number of classes $C$, number of samples $n_s$
\STATE $\mathcal{L}_0 \leftarrow \{(\vx_i, y_i)\}_{i=1}^j$ initial (balanced) labeled dataset
\STATE $\mathcal{U}_0 \leftarrow \{\vx_i\}_{i=j+1}^N$
\STATE $M_0 \leftarrow \text{Model trained on } \mathcal{L}_0$
\FOR{$k = 1,\ldots, K$}
\STATE $B_k \leftarrow \text{Info-NN-$m$} (M_{k-1}, \mathcal{L}_{k-1}, \mathcal{U}_{k-1}, b, C, n_s)$
\STATE $\mathcal{L}_k \leftarrow \mathcal{L}_{k-1} \cup \{(x_i,y_i) : x_i \in B_k\}$
\STATE $\mathcal{U}_k \leftarrow \mathcal{U}_{k-1} \backslash B_k$
\STATE $M_k \leftarrow \text{Model trained on } \mathcal{L}_k$
\ENDFOR\\
%\algorithmicoutput ~$M_K$
\end{algorithmic}
\end{algorithm}

\begin{algorithm}
\caption{Info-NN-$m$}
\label{alg: info_nn_cluster}
\begin{algorithmic} 
\REQUIRE Model $M$, labeled set $\mathcal{L}$, unlabeled set $\mathcal{U}$, batch size $b$, number of classes $C$, number of samples $n_s$
\STATE $\mZ_{\mathcal{L}}$ = Compute Embedding ($\mathcal{L}$)
\STATE $\mZ_{\mathcal{U}}$ = Compute Embedding ($\mathcal{U}$)
\STATE $Q \leftarrow \{\}$ (Set of candidate queries)
\FOR{$u \in \mZ_{\mathcal{U}}$}
\STATE $\text{NN}_u \leftarrow$ Top $m$ nearest neighbors 
\STATE $Q_u \leftarrow u \cup \text{NN}_u$
\STATE $Q \leftarrow Q \cup Q_u$
\ENDFOR
%\STATE $I \leftarrow \{\}$ (Set of mutual information values for the candidate queries)
%\FOR{$q \in Q$}
%\STATE Generate $n_s$ samples of $D_s \sim \mathcal{N}(D_q, \sigma^2)$
%\STATE $I_q \leftarrow H \left[ \underset{D \in D_s}{\E} \left( p(Y_q    \: | \:   D) \right) \right] - \underset{D \in D_s}{\E} \left( H \left[ p(Y_q    \: | \:   D) \right] \right)$
%\STATE $I \leftarrow I \cup I_q$
%\ENDFOR
\STATE $I \leftarrow $Info-NN-distances$(\mZ_{\mathcal{U}}, Q, n_s)$
\STATE $G(\mathcal{U}) \leftarrow$ Clustering ($\mathcal{U}, \mathcal{L}$)
\STATE $B \leftarrow$ unlabeled samples corresponding to top values of $I$ from every cluster\\
%\algorithmicoutput ~$B$
\end{algorithmic}
\end{algorithm}

\newpage
\subsection{Experimental details}

\paragraph{Computational infrastructure}

The experiments were performed on two desktop machines with the following configurations:
\begin{enumerate}[leftmargin=*]
    \item A 3.80GHz 16-Core Intel $i7-9800X$ CPU and an Nvidia Quadro RTX 5000 GPU
    \item A 2.10GHz 20-core Intel Xeon Gold 6230 CPU and four Nvidia Quadro RTX 6000 GPUs
\end{enumerate}

\paragraph{Datasets.}

Below are the details of the real world datasets used on classification experiments.
\begin{itemize}[leftmargin=*]
    \item MNIST \cite{lecun1998gradient} is a dataset of black and white images of handwritten digits belonging to 10 classes and consists 60,000 training samples and 10,000 test samples.
    \item CIFAR-10 \cite{krizhevsky2009learning} is a dataset consisting of colour images belonging to 10 classes with 50,000 training samples and 10,000 test samples.
    \item SVHN \cite{netzer2011reading} consists of digits (10 classes) from natural scene RGB images with 73,257 training samples and we use 10,000 samples for testing the accuracy of the learned models.
\end{itemize}

\paragraph{Baselines.}

The details of the baseline active labelling methods used are as follows.
\begin{itemize}[leftmargin=*]
    \item BatchBALD: Samples are selected according to the algorithm described in \cite{kirsch2019batchbald}. The algorithm uses Monte-Carlo (MC) sampling to compute joint probabilities of the different labelling configurations in a batch of samples which is very memory intensive. This requires the pool of unlabelled data to be sub-sampled in order for the computations to be feasible. The number of MC samples for the computations and the size of the pool set was determined by the memory associated with the GPUs. We use $10^3$ MC samples and the sizes of the pool set used were 20,000 for MNIST and 5000 for both CIFAR-10 and SVHN respectively. We would like to note here that we did not perform an extensive experimentation to determine an optimal configuration of the number of MC samples and size of the pool set but decided a configuration based on the settings that did not result in running out of GPU memory.
    \item K-Center: Optimal samples that achieve the desired coverage, based on the distances in the embedding space learned by the network, are selected.
    \item MaxEntropy: The top unlabeled samples with the maximum entropy, computed based on the class probabilities predicted by the model, are chosen.
    \item Random: A batch of samples is drawn at random from the pool for labelling.
\end{itemize}

\paragraph{Models and training methodology.}

% \paragraph{Supervised classification.}

In all the experiments, the models are trained from scratch at every active learning cycle. The performance reported is measured on a holdout test set comprising of 10,000 samples in all the experiments.

\paragraph{MNIST:} For experiments on the MNIST dataset, we use a model similar to the one used in \cite{kirsch2019batchbald}. Specifically, we use a CNN consisting of two convolutional blocks followed by two fully connected layers. The two convolutional blocks consist of $32$ and $64$ filters of kernel size $5$, each followed by layers of dropout, max-pooling and relu units. The two fully connected layers, of size 128 and 10 respectively, also have a dropout unit between them. We use a probability of 0.5 for all dropout units. 

The data inputs to the model are normalized and batch sizes of 64 and 1000 are used while training and testing respectively. We use the Adam optimizer with a learning rate of 0.001. Since the size of the labeled set used in these experiments is small compared to the entire dataset, we use early stopping to ensure that the model does not overfit to the training data. We use a validation set of size 100 consisting of 10 samples from every class selected at random and we stop training after 10 consecutive epochs of increasing validation loss. 

\paragraph{CIFAR-10 and SVHN:} For both the datasets, we use a ResNet-18 \cite{he2016deep} to conduct the experiments. While training, the data inputs are normalized along with augmentation techniques consisting of random cropping with an output size of 32 and a padding of 4 and random horizontal flipping. The model is trained for 250 epochs using the Adam optimizer with a learning rate of 0.001 in combination with the cosine annealing scheduler. A batch size of 128 is used for both training and testing.

% \newpage
\paragraph{Info-NN configuration.}
\paragraph{Inference hyperparameters:} The parameter $\mu$ is set equal to the maximum value of the inter-sample distances in the embedding space. The standard deviation for the normal distribution of distances is set as the standard deviation of all the distances in the embedding space and $1000$ samples from the distributions are used for inference. These values were found to work well in all the experiments and an extensive and a systemic search for these hyperparameters was not performed.

\paragraph{Clustering:} For MNIST, we initially use $K$-Means as the clustering technique and then switch to a $K$-NN based method where every unlabelled sample is grouped into one among the 10 classes based on the top $5$ nearest labelled samples. This works better for MNIST since we start with a very small amount of labelled data which makes $K$-NN based clustering not very effective at the beginning. We use $K$-Means for CIFAR-10 and SVHN datasets.

\subsection{Additional results}

\begin{figure}[t]
\captionsetup[sub]{justification=centering}
\begin{subfigure}{0.31\textwidth}
    \includegraphics[width=\textwidth]{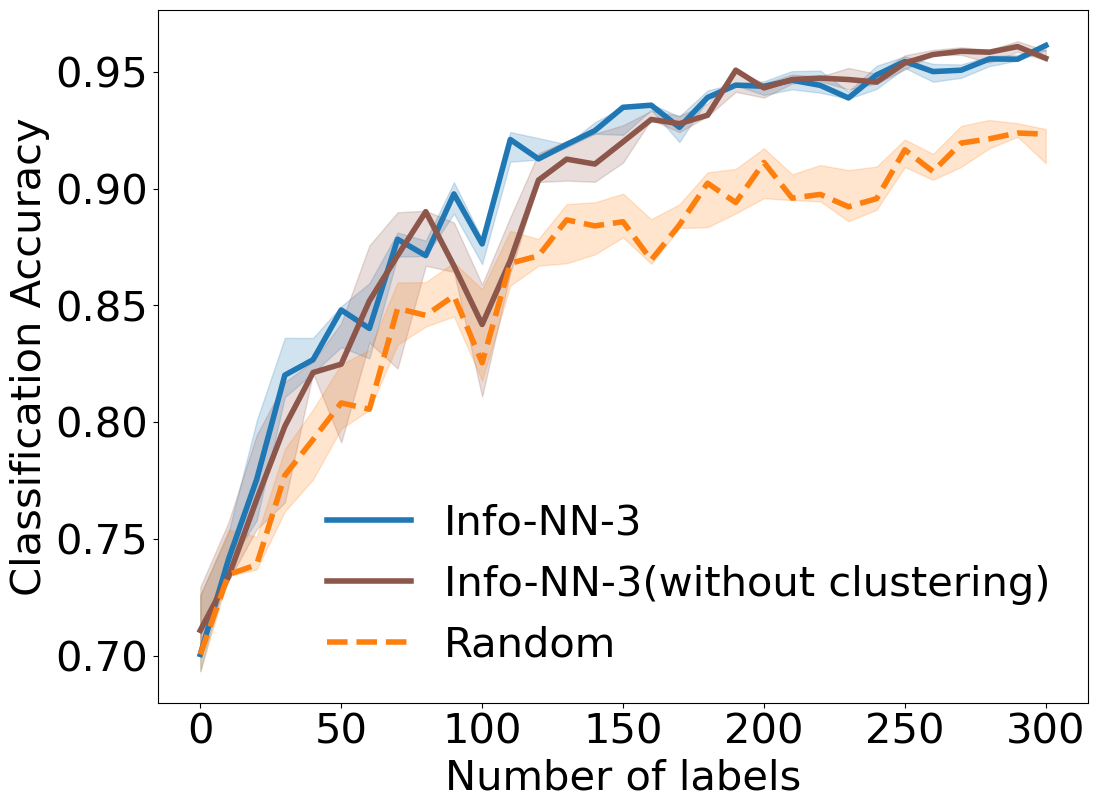}
    \label{fig:mnist_2_versions}
    \vspace{-1\baselineskip}
\end{subfigure}
\hspace*{\fill} % separation between the subfigures
\begin{subfigure}{0.31\textwidth}
    \includegraphics[width=\textwidth]{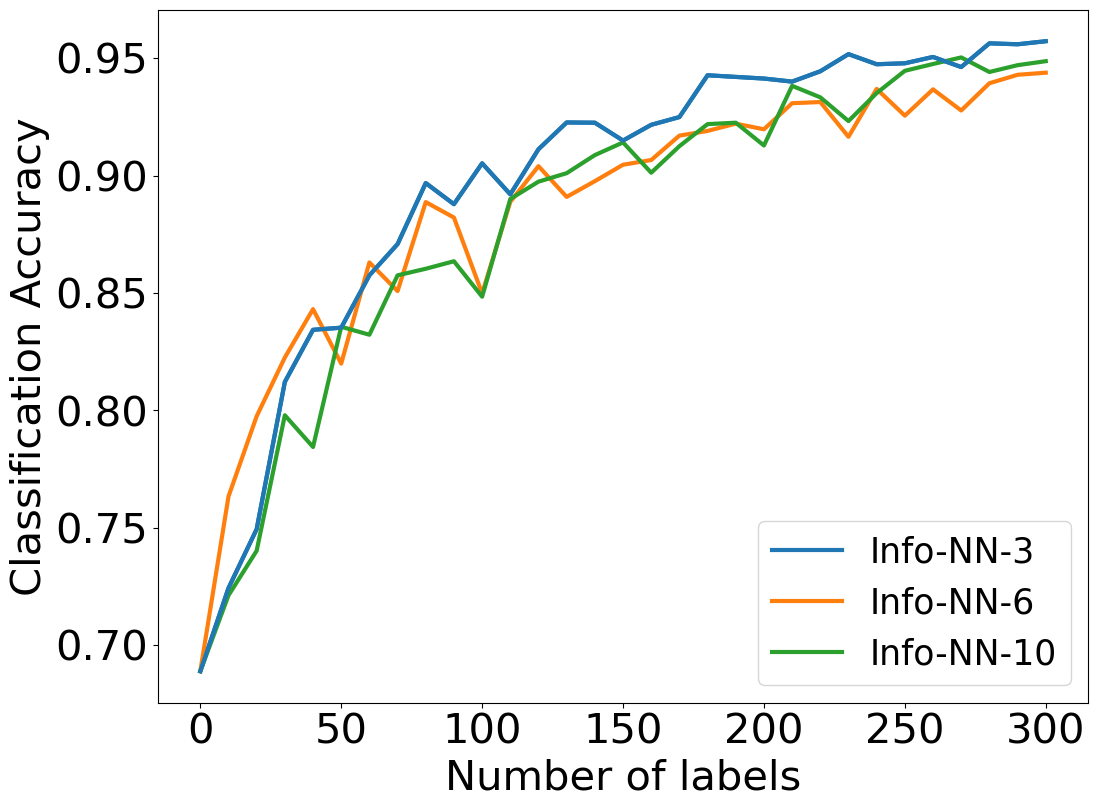}
    \label{fig:mnist_diff_query_lengths}
    \vspace{-1\baselineskip}
\end{subfigure}
\hspace*{\fill} % separation between the subfigures
\begin{subfigure}{0.31\textwidth}
    \includegraphics[width=\textwidth]{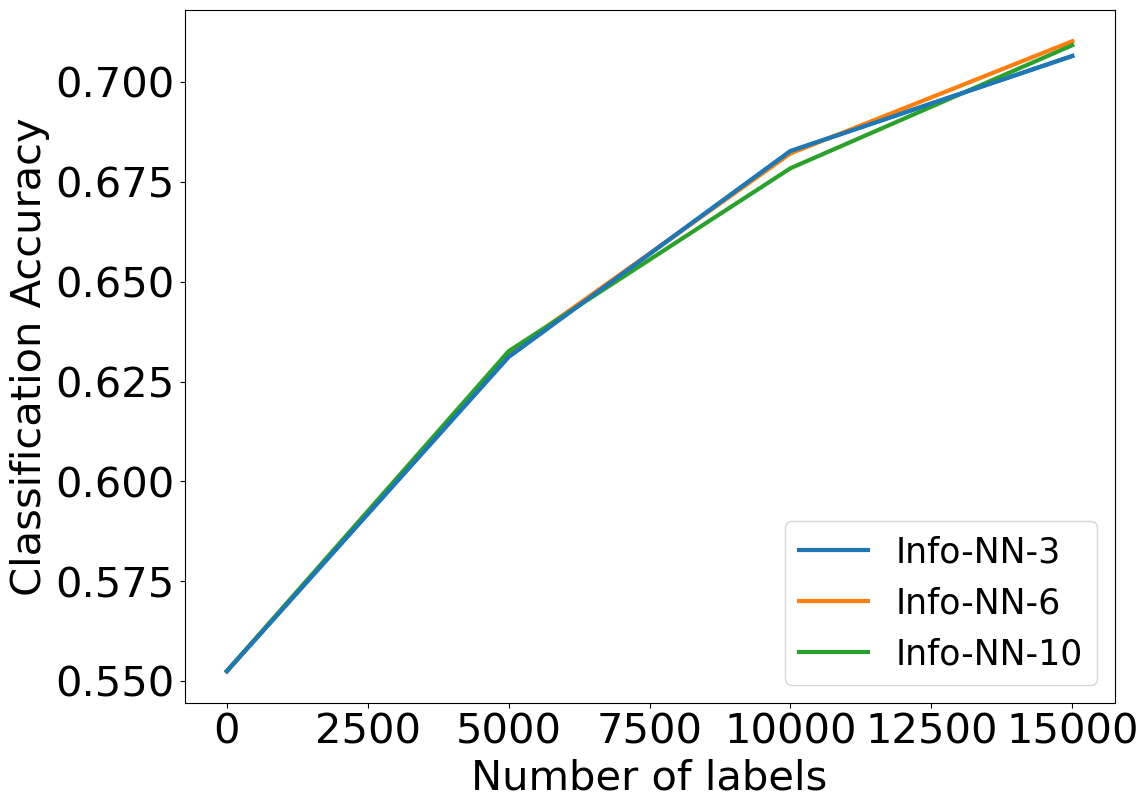}
    \label{fig:cifar10_diff_query_lengths}
    \vspace{-1\baselineskip}
\end{subfigure}
\caption{Active classification experiments: Comparison of the performances of Info-NN with and without clustering using a batch size of 3 on the MNIST dataset (left). Performance comparison between Info-NN queries of different lengths on MNIST (center) and CIFAR-10 (right) datasets.}
\label{fig:classification_more_results}
%\vspace{-2mm}
\end{figure}

\paragraph{Performance plots.}

We compare the performance of Info-NN with and without clustering (top $b$ samples are selected solely based on informativeness) on MNIST. The results are illustrated in Fig.~\ref{fig:classification_more_results} where we can observe the improved performance realized by Info-NN when combined with clustering. \\
\\
Also, we conducted experiments on MNIST and CIFAR-10 datasets to determine the optimal query length for Info-NN. In Fig.\ref{fig:classification_more_results}, we can observe that queries of length 3 resulted in the best performance on MNIST, significantly outperforming queries of longer lengths. On CIFAR-10, while all of them seem to exhibit a similar performance, queries of length 3 outperform the others consistently. Thus, we use queries of length 3 in all the experiments with supervised classification.

\newpage
\paragraph{Visualizations.}

\begin{figure}[t]
\captionsetup[sub]{justification=centering}
\begin{subfigure}{0.31\textwidth}
    \includegraphics[width=\textwidth]{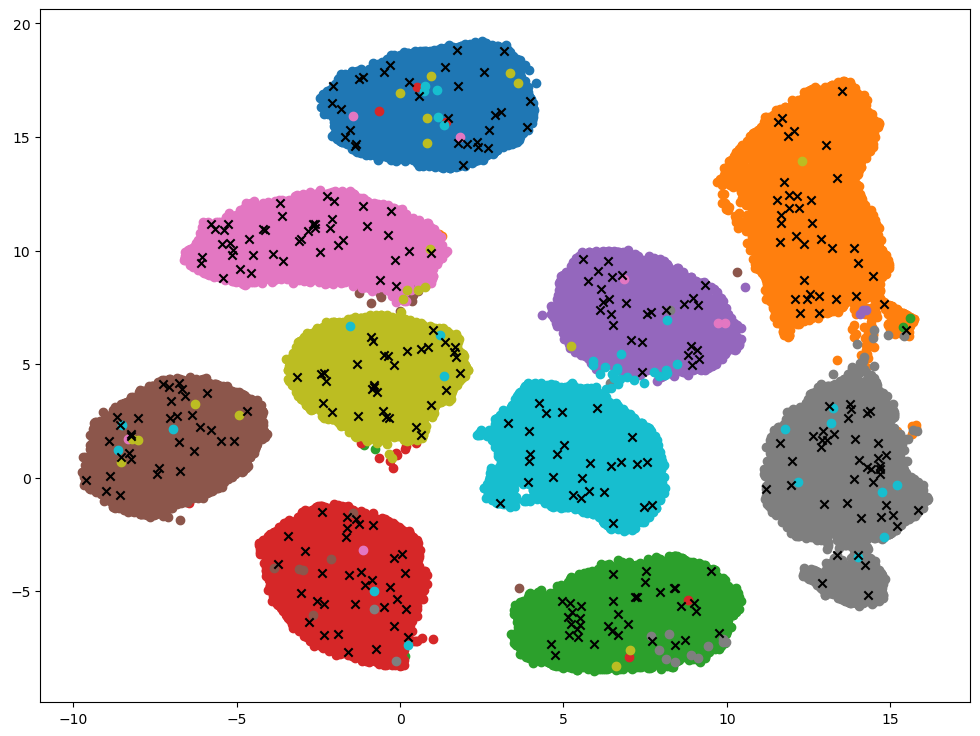}
    %\label{fig:mnist_info_nn}
    \vspace{-1\baselineskip}
    %\caption{}
\end{subfigure}
\hspace*{\fill} % separation between the subfigures
\begin{subfigure}{0.31\textwidth}
    \includegraphics[width=\textwidth]{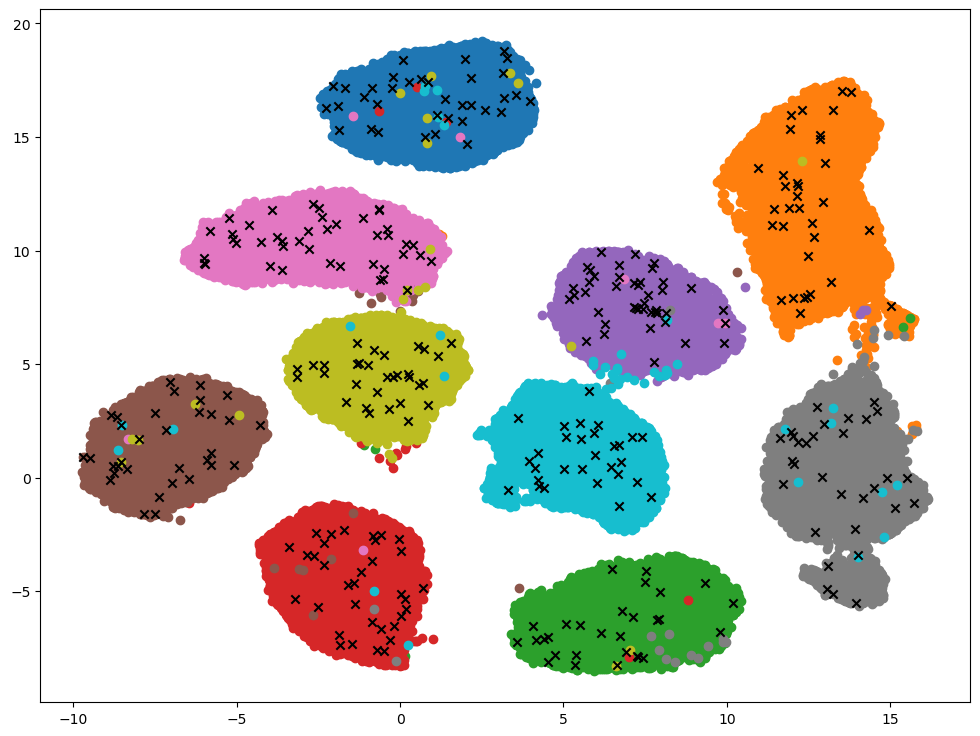}
    %\label{fig:mnist_baselines}
    \vspace{-1\baselineskip}
    %\caption{}
\end{subfigure}
\hspace*{\fill}
\begin{subfigure}{0.31\textwidth}
    \includegraphics[width=\textwidth]{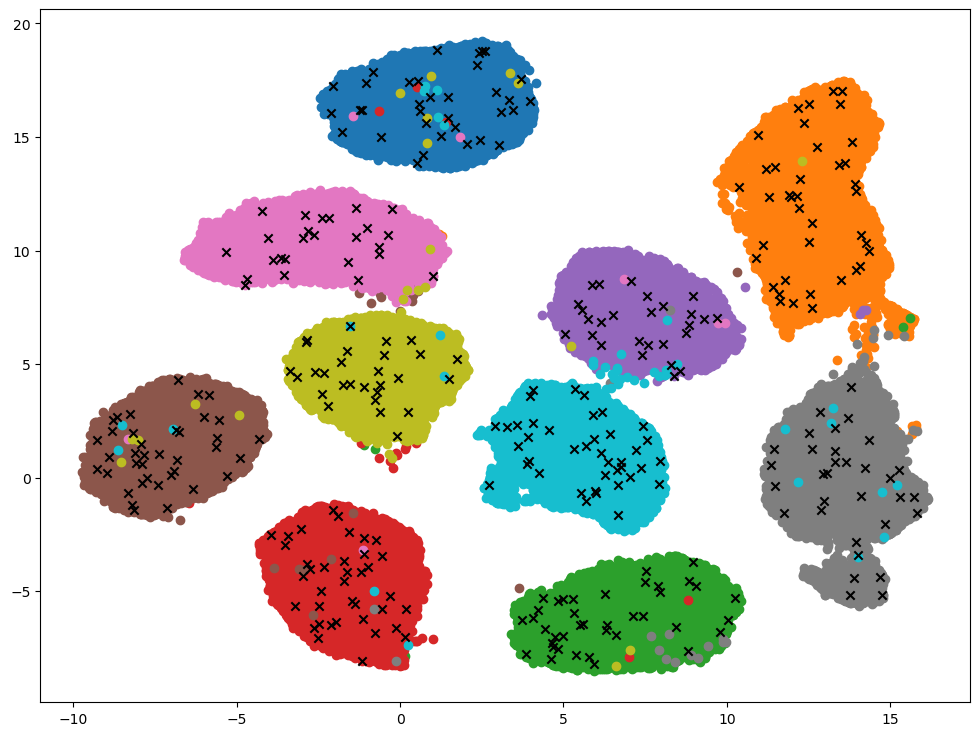}
    %\label{fig:cifar10_baselines}
    \vspace{-1\baselineskip}
    %\caption{}
\end{subfigure}
\caption{Visualization of samples selected on MNIST with MaxEntropy (left), Info-NN-3 (center) and K-Center (right) querying strategies, generated using UMAP \cite{mcinnes2018umap}. Each of the blobs correspond to one among the 10 classes and the samples selected are indicated by black crosses.}
\label{fig:samples_selected}
%\vspace{-2mm}
\end{figure}

The samples selected by different active methods are illustrated in  Fig.~\ref{fig:samples_selected}. We can observe that MaxEntropy tends to select redundant informative samples indicated by clusters of black crosses and K-Center selects samples to ensure diversity indicated by the more distributed placement of the selected samples. In the case of Info-NN, we see a combination of clustered and distributed samples likely selecting both informative and diverse samples.